\documentclass[10pt,twocolumn,letterpaper]{article}

\usepackage{iccv}
\usepackage{times}
\usepackage{epsfig}
\usepackage{graphicx}
\usepackage{amsmath}
\usepackage{amssymb}
\usepackage{booktabs}
\usepackage{subcaption}
\usepackage{booktabs}
\DeclareMathOperator*{\argmax}{arg\,max}

\usepackage{fixltx2e}
\usepackage{enumitem}

\usepackage[pagebackref=true,breaklinks=true,letterpaper=true,colorlinks,bookmarks=false]{hyperref}

\iccvfinalcopy 

\def\iccvPaperID{6813} 

\ificcvfinal\fi

\begin{document}

\title{FaceCLIPNeRF: Text-driven 3D Face Manipulation using Deformable \\ Neural Radiance Fields}

\newcommand*{\affaddr}[1]{#1}
\newcommand*{\affmark}[1][*]{\textsuperscript{#1}}
\newcommand*{\email}[1]{\texttt{#1}}
\author{
Sungwon Hwang\affmark[1] \quad Junha Hyung\affmark[1]\quad Daejin Kim\affmark[2]\quad Min-Jung Kim\affmark[1]\quad Jaegul Choo\affmark[1]\\
\\
\affaddr{\affmark[1]KAIST}\quad \affaddr{\affmark[2]Scatter Lab}\\
\small\email{\{shwang.14, sharpeeee, emjay73, jchoo\}@kaist.ac.kr, daejin@scatterlab.co.kr}
}

\twocolumn[{
\maketitle
\vspace{-0.5cm}
\begin{center}
    \centering
    \captionsetup{type=figure}
    \includegraphics[width=\linewidth]{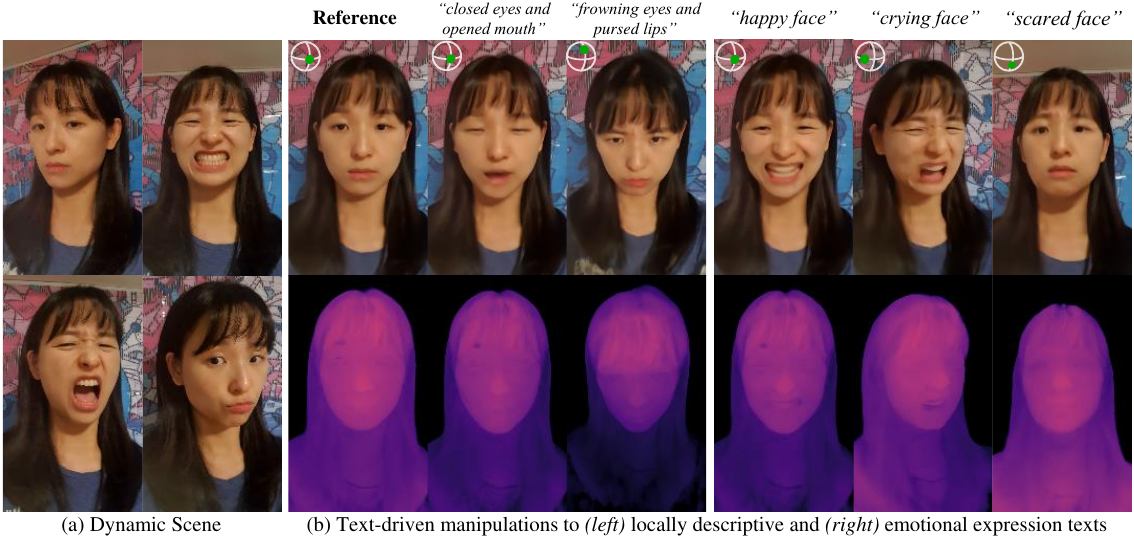}
    \captionof{figure}{\textbf{FaceCLIPNeRF} reconstructs a video of a dynamic scene of a face, and conducts face manipulation using texts only. Manipulated faces and their depths in top and bottom rows in (b), respectively, are rendered from novel views.}
    \label{fig:front}
\end{center}
}]

\ificcvfinal\thispagestyle{empty}\fi

\begin{abstract}
\vspace{-0.4cm}
As recent advances in Neural Radiance Fields (NeRF) have enabled high-fidelity 3D face reconstruction and novel view synthesis, its manipulation also became an essential task in 3D vision. However, existing manipulation methods require extensive human labor, such as a user-provided semantic mask and manual attribute search unsuitable for non-expert users. Instead, our approach is designed to require a single text to manipulate a face reconstructed with NeRF. To do so, we first train a scene manipulator, a latent code-conditional deformable NeRF, over a dynamic scene to control a face deformation using the latent code. However, representing a scene deformation with a single latent code is unfavorable for compositing local deformations observed in different instances. As so, our proposed Position-conditional Anchor Compositor (PAC) learns to represent a manipulated scene with spatially varying latent codes. Their renderings with the scene manipulator are then optimized to yield high cosine similarity to a target text in CLIP embedding space for text-driven manipulation. To the best of our knowledge, our approach is the first to address the text-driven manipulation of a face reconstructed with NeRF. Extensive results, comparisons, and ablation studies demonstrate the effectiveness of our approach.

\end{abstract}
\vspace{-0.5cm}
\section{Introduction}
\label{sec:intro}
Easy manipulation of 3D face representation is an essential aspect of advancements in 3D digital human contents\cite{sharma20223d}. Though Neural Radiance Field\cite{mildenhall2020nerf} (NeRF) made a big step forward in a 3D scene reconstruction, many of its manipulative methods targets color\cite{fan2022unified, sun2022fenerf} or rigid geometry \cite{yuan2022nerf, lazova2022control, yang2021objectnerf, kobayashi2022distilledfeaturefields} manipulations, which are inappropriate for detailed facial expression editing tasks. While a recent work proposed a regionally controllable face editing method \cite{kania2022conerf}, it requires an exhaustive process of collecting user-annotated masks of face parts from curated training frames, followed by manual attribute control to achieve a desired manipulation. Face-specific implicit representation methods \cite{gafni2021dynamic, zheng2022avatar} utilize parameters of morphable face models \cite{thies2016face2face} as priors to encode observed facial expressions with high fidelity. However, their manipulations are not only done manually but also require extensive training sets of approximately 6000 frames that cover various facial expressions, which are laborious in both data collection and manipulation phases. On the contrary, our approach only uses a single text to conduct facial manipulations in NeRF, and trains over a dynamic portrait video with approximately 300 training frames that include a few types of facial deformation examples as in Fig. \ref{fig:front}\textcolor{red}{a}. 

\begin{figure}[t]
  \begin{subfigure}{\linewidth}
  \includegraphics[width=\linewidth]{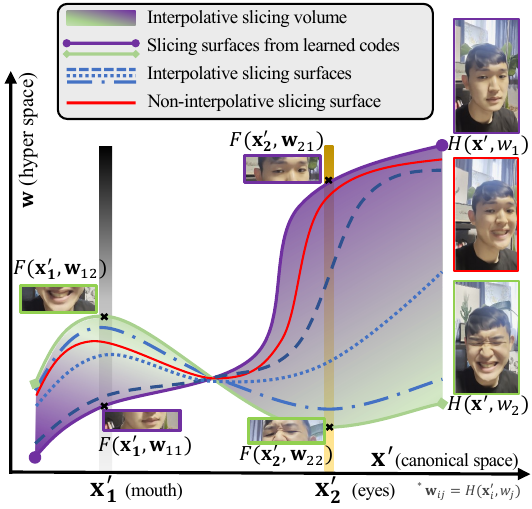}
  \vspace{-0.7cm}
  \caption{}
  \label{fig:2.a}
  \end{subfigure}

  \begin{subfigure}{0.208\linewidth}
      \includegraphics[width=\linewidth]{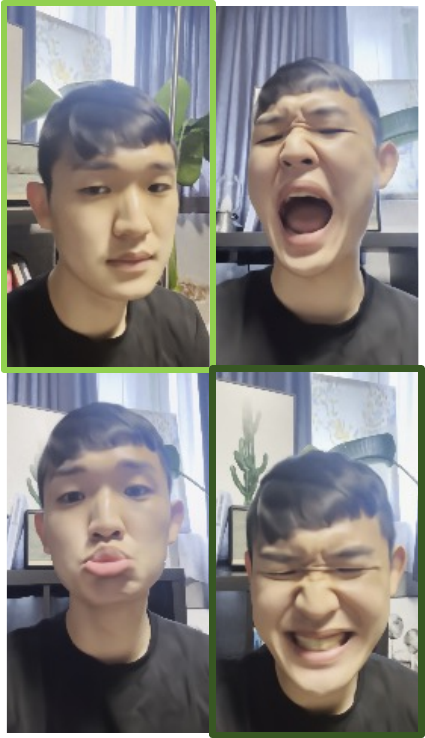}
      \caption{}
      \label{fig:2.b}
  \end{subfigure}
  \begin{subfigure}{0.792\linewidth}
      \includegraphics[width=\linewidth]{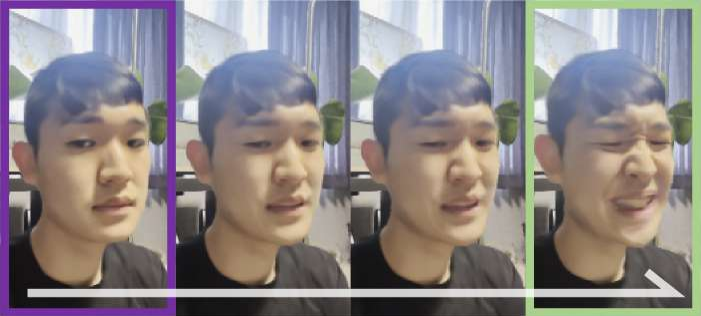}
      \caption{}
      \label{fig:2.c}
  \end{subfigure}
  \vspace{-0.6cm}
  \caption{(a) Illustration of linked local attribute problem in hyper space. Expressing scene deformation with per-scene latent code cannot compose local facial deformation observed in different instances. (b) Types of facial deformations observed during scene manipulator training. (c) Renderings of interpolated latent codes with a scene manipulator.
  }
  \label{fig:2}
  \vspace{-0.4cm}
\end{figure}

In order to control a face deformation, our method first learns and separates observed deformations from a canonical space leveraging HyperNeRF\cite{park2021hypernerf}. Specifically, per-frame deformation latent codes and a shared latent code-conditional implicit scene network are trained over the training frames. Our key insight is to represent the deformations of a scene with multiple, spatially-varying latent codes for manipulation tasks. The insight originates from the shortcomings of na\"ively adopting the formulations of HyperNeRF to manipulation tasks, which is to search for a single latent code that represents a desired face deformation. For instance, a facial expression that requires a combination of local deformations observed in different instances is not expressible with a single latent code. In this work, we define such a problem as \emph{``linked local attribute problem"} and address this issue by representing a manipulated scene with spatially varying latent codes. As a result, our manipulation could express a combination of locally observed deformations as seen from the image rendering highlighted with red boundary in Fig. \ref{fig:2.a}.

To this end, we first summarize all observed deformations as a set of anchor codes and let MLP learn to compose the anchor codes to yield multiple, position-conditional latent codes. The reflectivity of the latent codes on visual attributes of a target text is then achieved by optimizing the rendered images of the latent codes to be close to a target text in CLIP\cite{radford2021learning} embedding space. In summary, our work makes the following contributions:
\vspace{-0.1cm}
\begin{itemize}
    \item Proposal of a text-driven manipulation pipeline of a face reconstructed with NeRF.
    \item Design of a manipulation network that learns to represent a scene with spatially varying latent codes. 
    \item First to conduct text-driven manipulation of a face reconstructed with NeRF to the best of our knowledge.
\end{itemize}
\vspace{-0.25cm}

\section{Related Works}
\label{sec:relworks}


\paragraph{NeRF and Deformable NeRF}
Given multiple images taken from different views of a target scene, NeRF\cite{mildenhall2020nerf} synthesizes realistic novel view images with high fidelity by using an implicit volumetric scene function and volumetric rendering scheme\cite{kajiya1984ray}, which inspired many follow-ups \cite{barron2021mip, tancik2021learned, mildenhall2022nerf, verbin2022refnerf, yu2021plenoctrees}. As NeRF assumes a static scene, recent works \cite{park2021nerfies, park2021hypernerf, pumarola2020d, li2022neural} propose methods to encode dynamic scenes of interest. The common scheme of the works is to train a latent code per training frame and a single latent-conditional NeRF model shared by all trained latent codes to handle scene deformations. Our work builds on this design choice to learn and separate the observed deformations from a canonical space, yet overcome its limitation during the manipulation stage by representing a manipulated scene with spatially varying latent codes. 

\vspace{-0.3cm}
\paragraph{Text-driven 3D Generation and Manipulation}
Many works have used text for images or 3D manipulation\cite{wang2022clip,jain2022zero,poole2022dreamfusion, jetchev2021clipmatrix, sanghi2022clip, jain2021dreamfields}.
CLIP-NeRF\cite{wang2022clip} proposed a disentangled conditional NeRF architecture in a generative formulation supervised by text embedding in CLIP\cite{radford2021learning} space, and conducted text-and-exemplar driven editing over shape and appearance of an object.
Dreamfields \cite{jain2022zero} performed generative text-to-3D synthesis by supervising its generations in CLIP embedding space to a generation text.
We extend from these lines of research to initiate CLIP-driven manipulation of face reconstructed with NeRF.
\vspace{-0.3cm}
\paragraph{NeRF Manipulations}
Among many works that studied NeRF manipulations\cite{liu2021editing, yuan2022nerf,thies2016face2face,kania2022conerf,sun2022fenerf,sun2022ide,hong2022headnerf,zhuang2021mofanerf,lazova2022control},
EditNeRF\cite{liu2021editing} train conditional NeRF on a shape category to learn implicit semantics of the shape parts without explicit supervision. Then, its manipulation process propagates user-provided scribbles to appropriate object regions for editing.
NeRF-Editing\cite{yuan2022nerf} extracts mesh from trained NeRF and lets the user perform the mesh deformation. A novel view of the edited scene can be synthesized without re-training the network by bending corresponding rays.
CoNeRF\cite{kania2022conerf} trains controllable neural radiance fields using user-provided mask annotations of facial regions so that the user can control desired attributes within the region.
However, such methods require laborious annotations and manual editing processes, whereas our method requires only a single text for detailed manipulation of faces.
\vspace{-0.3cm}
\paragraph{Neural Face Models}

Several works\cite{yenamandra2021i3dmm, ramon2021h3d, zheng2022avatar} built 3D facial models using neural implicit shape representation. Of the works, i3DMM\cite{yenamandra2021i3dmm} disentangles face identity, hairstyle, and expression, making decoupled components to be manually editable. 
Face representation works based on NeRF have also been exploited\cite{wang2021learning, thies2016face2face, zheng2022avatar}. 
Wang \textit{et al.}\cite{wang2021learning} proposed compositional 3D representation for photo-realistic rendering of a human face, yet requires guidance images to extract implicitly controllable codes for facial expression manipulation. 
NerFACE\cite{thies2016face2face} and IMavatar\cite{zheng2022avatar} model the appearance and dynamics of a human face using learned 3D Morphable Model\cite{blanz1999morphable} parameters as priors to achieve controllability over pose and expressions. However, the methods require a large number of training frames that cover many facial expression examples and manual adjustment of the priors for manipulation tasks. 

\section{Preliminaries}

\subsection{NeRF}
NeRF \cite{mildenhall2020nerf} is an implicit representation of geometry and color of a space using MLP. Specifically, given a point coordinate $\textbf{x} = (x, y, z)$ and a viewing direction $\textbf{d}$, an MLP function $\mathcal{F}$ is trained to yield density and color of the point as $(\textbf{c}, \sigma) = \mathcal{F}(\textbf{x}, \textbf{d})$. $M$ number of points are sampled along a ray $\textbf{r} = \textbf{o} + t\textbf{d}$ using distances, $\{t_{i}\}_{i=0}^{M}$, that are collected from stratified sampling method. $F$ predicts color and density of each point, all of which are then rendered to predict pixel color of the ray from which it was originated as


\begin{equation}
    \label{eq:nerfrender}
    \hat{C}(\textbf{r}) = \sum_{i=1}^{M} T_{i}(1-\text{exp}(-\sigma_{i}\delta_{i}))\textbf{c}_{i}, 
\end{equation}

\noindent where $\delta_{i} = t_{i+1} - t_{i}$, and $T_{i} = \text{exp}(-\sum_{j=1}^{i-1}\sigma_{j}\delta_{j})$ is an accumulated transmittance. $\mathcal{F}$ is then trained to minimize the rendering loss supervised with correspondingly known pixel colors. 

\subsection{HyperNeRF}
\label{section: hypernerf}
Unlike NeRF that is designed for a static scene, HyperNeRF \cite{park2021hypernerf} is able to encode highly dynamic scenes with large topological variations. Its key idea is to project points to canonical hyperspace for interpretation. Specifically, given a latent code $w$, a spatial deformation field $T$ maps a point to a canonical space, and a slicing surface field $H$ determines the interpretation of the point for a template NeRF $F$. Specifically,  
\begin{align}
    \textbf{x}' = T(\textbf{x}, w) \label{eq:hypert}, \ \ \ \ \ \ \ \\
    \textbf{w} = H(\textbf{x}, w) \label{eq:hyperh}, \ \ \ \ \ \ \ \\
    (\textbf{c}, \sigma) = F(\textbf{x}', \textbf{w}, \textbf{d}),\label{eq:hyperf}
\end{align}

\noindent where $w \leftarrow w_n \in \{w_{1} \cdots w_{N}\} = W$ is a trainable per-frame latent code that corresponds to each $N$ number of training frames. Then, the rendering loss is finally defined as

\begin{equation}
    \mathcal{L}_{c} = \sum_{\substack{n \in \{1 \cdots N\}, \\ \textbf{r}^{n} \in \mathcal{R}^{n}}}||C_{n}(\textbf{r}^{n}) - \hat{C}_{n}(\textbf{r}^{n})||_{2}^{2},
\end{equation}

\noindent where $C_{n}(\textbf{r}^{n})$ is ground truth color at $n$-th training frame of a ray $\textbf{r}^{n}$ and $\mathcal{R}^{n}$ is a set of rays from $n$-th camera. Note that $(\textbf{x}', \textbf{w})$ and $H(\textbf{x}, w)$ are often referred to \textit{canonical hyperspace} and \textit{slicing surface}, since $\textbf{x}'$ can be interpreted differently for different $w$ as illustrated in Fig.~\ref{fig:2.a}.



\section{Proposed Method}
We aim to manipulate a face reconstructed with NeRF given a target text that represents a desired facial expressions for manipulation (e.g., ``\textit{crying face}'', ``\textit{wink eyes and smiling mouth}''). To this end, our proposed method first trains a scene manipulator, a latent code-conditional neural field that controls facial deformations using its latent code (\S\ref{subsec:4.1}). Then, we elaborate over the pipeline to utilize a target text for manipulation (\S\ref{subsec:4.2}), followed by proposing an MLP network that learns to appropriately use the learned deformations and the scene manipulator to render scenes with faces that reflect the attributes of target texts (\S\ref{subsec:4.3}).

\begin{figure*}
  \includegraphics[width=\linewidth]{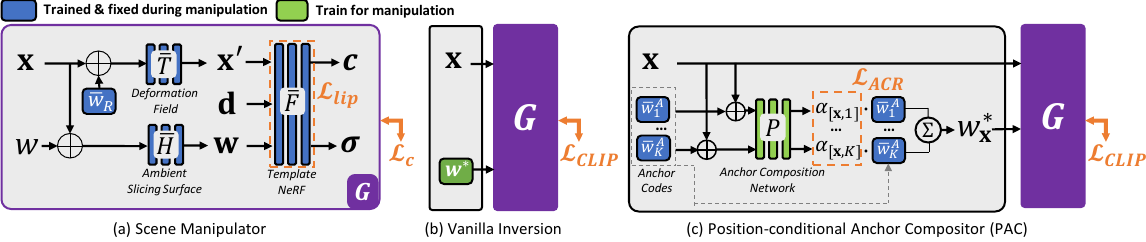}
  \caption{(a) Network structure of scene manipulator $G$. (b) Vanilla inversion method for manipulation. (c) Position-conditional Anchor Compositor (PAC) for manipulation.}
  \label{fig:3}
  \vspace{-0.3cm}
\end{figure*}

\subsection{Scene Manipulator}
\label{subsec:4.1}



First, we construct a scene manipulator using HyperNeRF\cite{park2021hypernerf} so that deformations of a scene can be controlled by fixing the parameters of the scene manipulator and manipulating its latent code. Specifically, we train a dynamic scene of interest with a network formulated as Eq.\eqref{eq:hyperf} following \cite{park2021hypernerf}, after which we freeze the trained parameters of $T$, $H$, $F$, and $W$ and use $w$ as a manipulation handle. In addition, we empirically found that the deformation network $T$ tends to learn rigid deformations, such as head pose, while slicing surface field $H$ learns non-rigid and detailed deformations, such as shapes of mouth and eyes. As so, we select and fix a trained latent code for $T$ and only manipulate a latent code fed to $H$. In summary, as illustrated in Fig.~\ref{fig:3}(a), our latent code-conditional scene manipulator $G$ is defined as

\begin{equation}
    \label{eq:hyperfix}
    G(\textbf{x}, \textbf{d}, w) := \bar{F}(\bar{T}(\textbf{x}, \bar{w}_R), \bar{H}(\textbf{x}, w), \textbf{d}),
\end{equation}

\noindent where $\bar{\cdot}$ represents that the parameters are trained and fixed for manipulation, and $\bar{w}_{R}$ is a fixed latent code of the desired head pose chosen from a set of learned latent codes $\bar{W}$. In the supplementary material, we report further experimental results and discussions over head pose controllability of $\bar{w}_{R}$.  

\vspace{-0.25cm}
\paragraph{Lipschitz MLP} Since $G$ is only trained to be conditioned over a limited set of trainable latent codes $W$, a subspace of $w$ outside the learned latent codes that yields plausible deformations needs to be formulated to maximize the expressibility of $G$ for manipulation. Meanwhile, HyperNeRF was shown to moderately render images from latent codes linearly interpolated from two learned latent codes. Thus, a valid latent subspace $\mathcal{W}$ can be formulated to include not only the learned latent codes but codes linearly interpolated between any two learned latent codes as well. Specifically, 

\vspace{-0.05cm}
\begin{equation}
    \label{eq:2}
    \begin{aligned}
        \mathcal{W} \supset \{\gamma * \bar{w}_{i} + (1 - \gamma) * \bar{w}_{j} \ | \  \ \bar{w}_{i}, \bar{w}_{j} \in \bar{W}, \\
        \ 0 \leq \gamma \leq 1 \}.
    \end{aligned}
\end{equation}


However, we learned that the fidelity of images from interpolated latent codes needs to be higher to be leveraged for manipulation. As so, we regularize the MLPs of the scene manipulator to be more Lipschitz continuous during its training phase. Note that Lipschitz bound of a neural network with $L$ number of layers and piece-wise linear functions such as ReLU can be approximated as $c = \prod_{i=1}^{L} \| \text{W}^{i} \|_{p}$ \cite{liu2022learning, yoshida2017spectral}, where $\text{W}^{i}$ is an MLP weight at $i$-th layer. 
Since a function $f$ that is $c$-Lipschitz has the property

\begin{equation}
    \|f(w_1) - f(w_2) \|_{p} \leq c \|w_1 - w_2 \|_{p},
\end{equation}

\noindent successful regularization of $c$ would make smaller differences between outputs of adjacent latent codes, which induce interpolated deformations to be more visually natural. As so, we follow \cite{liu2022learning} and regularize trainable matrix at $l$-th layer of $F$ by introducing extra trainable parameters $c^{l}$ as

\vspace{-0.3cm}

\begin{equation}
    y^{l} = \sigma(\hat{\text{W}}^{l}x + b^{l}), \ \hat{\text{W}}^{l}_j = \text{W}^{l}_j \cdot \text{min}(1, \frac{softplus(c^{l})}{ \| \text{W}^{l}_j \|_{\infty}}), 
\end{equation}

\noindent where $\text{W}^{l}_j$ is the $j$-th row of a trainable matrix at $l$-th layer $\text{W}^{l}$, and $\| \cdot \|_{\infty}$ is matrix $\infty$-norm. Trinable Lipschitz constants from the layers are then minimized via gradient-based optimization with loss function defined as
\vspace{-0.15cm}
\begin{equation}
    \mathcal{L}_{lip} = \prod_{l=1}^{L}softplus(c^{l}).
\end{equation}


In summary, networks in Eq. \eqref{eq:hyperf} are trained to retrieve $\bar{F}$, $\bar{T}$, $\bar{H}$, and $\bar{W}$ using our scene manipulator objective function
\vspace{-0.05cm}
\begin{equation}
    \mathcal{L}_{SM} = \lambda_{c}\mathcal{L}_{c} + \lambda_{lip}\mathcal{L}_{lip},
\end{equation}

\noindent where $\lambda_{c}$ and $\lambda_{lip}$ are hyper-parameters.

\subsection{Text-driven Manipulation} 
\label{subsec:4.2}

Given a trained scene manipulator $G$, one manipulation method is to find a single optimal latent code $w$ whose rendered image using $G$ yields the highest cosine similarity with a target text in CLIP\cite{radford2021learning} embedding space, so that the manipulated images can reflect the visual attributes of a target text. Specifically, given images rendered with $G$ and $w$ at a set of valid camera poses $[R|t]$ as $\mathcal{I}^{G, w}_{[R|t]}$ and a target text for manipulation $p$, the goal of the method is to solve the following problem:
\vspace{-0.02cm}
\begin{equation}
    \label{eq:1}
    w^{*} = \argmax_{w} D_{\text{CLIP}}(\mathcal{I}^{G, w}_{[R|t]}, p), 
\end{equation}
 
\noindent where $D_{\text{CLIP}}$ measures the cosine similarity of features between rendered images and a target text extracted from pre-trained CLIP model. 

As illustrated in Fig.~\ref{fig:3}\textcolor{red}{b}, a straightforward vanilla approach to find an optimal latent embedding $w^{*}$ is inversion, a gradient-based optimization of $w$ that maximizes Eq.(\ref{eq:1}) by defining a loss function as $\mathcal{L}_{CLIP} = 1 - D_{\text{CLIP}}(\mathcal{I}^{G, w}_{[R|t]}, p)$. However, we show that this method is sub-optimal by showing that it inevitably suffers from what we define as a \textit{linked local attributes} problem, which we then solve with our proposed method. 

\vspace{-0.3cm}
\paragraph{Linked local attribute problem} Solutions from the vanilla inversion method are confined to represent deformations equivalent to those from $\mathcal{W}$. However, $\mathcal{W}$ cannot represent all possible combinations of locally observed deformations, as interpolations between two learned latent codes, which essentially comprise $\mathcal{W}$, cause facial attributes in different locations to change simultaneously. For example, consider a scene with deformations in Fig.~\ref{fig:2.b} and renderings of interpolations between two learned latent codes in Fig.~\ref{fig:2.c}. Not surprisingly, neither the learned latent codes nor the interpolated codes can express opened eyes with opened mouth or closed eyes with a closed mouth. Similar experiments can be done with any pair of learned latent codes and their interpolations to make the same conclusion. 

We may approach this problem from the slicing surface perspective of canonical hyperspace introduced in Sec. \ref{section: hypernerf}. As in Fig.~\ref{fig:2.a}, hyperspace allows only one latent code to represent an instance of a slicing surface representing a global deformation of all spatial locations. Such representation causes a change in one type of deformation in one location to entail the same degree of change to another type of deformation in different locations during interpolation.

Our method is motivated by the observation and is therefore designed to allow different position $\textbf{x}$ to be expressed with different latent codes to solve the linked local attribute problem.

\subsection{Position-conditional Anchor Compositor}
\label{subsec:4.3}

For that matter, Position-conditional Anchor Compositor (PAC) is proposed to grant our manipulation pipeline the freedom to learn appropriate latent codes for different spatial positions. 

Specifically, we define anchor codes
$\{\bar{w}^{A}_{1}, \cdots \bar{w}^{A}_{K}\} = \bar{W}^{A} \subset \bar{W}$ ,
a subset of learned latent codes where each represent different types of observed facial deformations, to set up a validly explorable latent space as a prior. We retrieve anchor codes by extracting facial expression parameters using DECA\cite{DECA:Siggraph2021} from images rendered from all codes in $\bar{W}$ over a fixed camera pose. Then, we cluster the extracted expression parameters using DBSCAN\cite{ester1996density} and select the latent code corresponding to the expression parameter closest to the mean for each cluster. For instance, we may get $K = 4$ anchor codes in the case of the example scenes in Fig.~\ref{fig:front}\textcolor{red}{a} and Fig.~\ref{fig:2.b}. 

\begin{figure}
  \includegraphics[width=\linewidth]{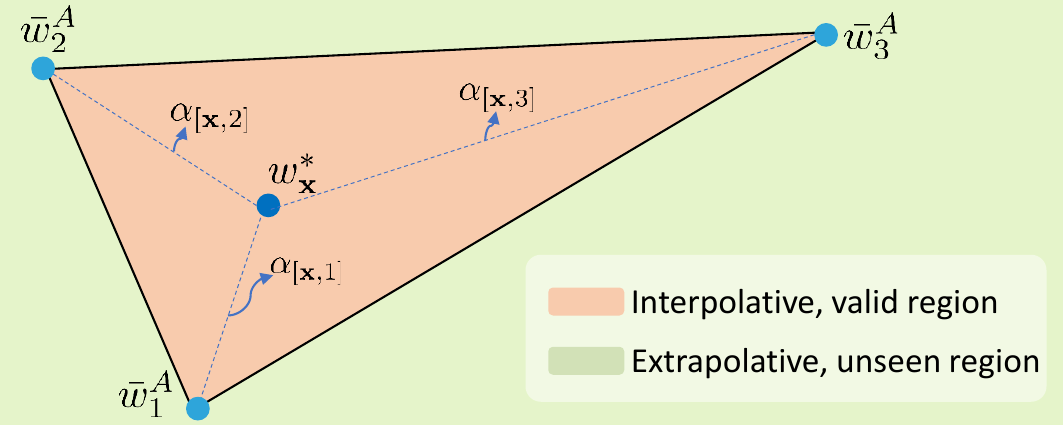}
  \caption{Illustration of barycentric interpolation of latent codes for validly expressive regions when $K=3$.}
  \label{fig:bary}
  \vspace{-0.3cm}
\end{figure}

Then for every spatial location, a position-conditional MLP yields appropriate latent codes by learning to compose these anchor codes. By doing so, a manipulated scene can be implicitly represented with multiple, point-wise latent codes. Specifically, the anchor composition network $P: \mathbb{R}^{(3+d_{w})} \rightarrow \mathbb{R}^{1} $ learns to yield $w^{*}_\textbf{x}$ for every spatial position $\textbf{x}$ via barycentric interpolation\cite{hormann2014barycentric} of anchors as
\vspace{-0.05cm}

\begin{equation}
    \hat{\alpha}_{[\textbf{x},k]} = P(\textbf{x} \oplus \bar{w}^A_k), \ \ w^{*}_{\textbf{x}} = \sum_k \sigma_{k}(\hat{\alpha}_{[\textbf{x},k]}) \bar{w}^{A}_k,
\end{equation}
\vspace{-0.2cm}




\noindent where $d_w$ is the dimension of a latent code, $\oplus$ is concatenation, and $\sigma_k$ is softmax activation along $k$ network outputs. Also, denote $\alpha_{[\textbf{x},k]} = \sigma_{k}(\hat{\alpha}_{[\textbf{x},k]})$ as anchor composition ratio (ACR) for ease of notation. As in the illustrative example in Fig.~\ref{fig:bary}, the key of the design is to prevent the composited code from diverging to extrapolative region of the latent. Thus, barycentric interpolation defines a safe bound of composited latent code for visually natural renderings.





Finally, a set of points that are sampled from rays projected at valid camera poses and their corresponding set of latent codes $[w^{*}_{\textbf{x}}]$ are queried by $G$, whose outputs are rendered as images to be supervised in CLIP embedding space for manipulation as

\vspace{-0.3cm}
\begin{equation}
\mathcal{L}_{CLIP} = 1 - D_{\text{CLIP}}(\mathcal{I}^{G, [w^{*}_{\textbf{x}}]}_{[R|t]}, p),
\end{equation}





\vspace{-0.3cm}
\paragraph{Total variation loss on anchor composition ratio} As, the point-wise expressibility of PAC allows adjacent latent codes to vary without mutual constraints, $P$ is regularized with total variation (TV) loss. Smoother ACR fields allows similar latent embeddings to cover certain facial positions to yield more naturally rendered images. Specifically, $\alpha_{[\textbf{x}, k]}$ is rendered to valid camera planes using the rendering equation in Eq. \eqref{eq:nerfrender} for regularization. Given a ray $\textbf{r}_{uv}(t) = \textbf{o} + t \textbf{d}_{uv}$, ACR can be rendered for each anchor $k$ at an image pixel located at $(u, v)$ of a camera plane, and regularized with TV loss as
\vspace{-0.1cm}
\begin{gather}
    \label{eq:alpha}
    \tilde{\alpha}_{kuv} = \sum_{i=1}^{M} T_{i}(1-\text{exp}(-\sigma_{i}\delta_{i}))\alpha_{[\textbf{r}_{uv}(t_{i}), k]}, \\
    \mathcal{L}_{ACR} = \sum_{k,u,v}  \| \tilde{\alpha}_{k(u+1)v} - \tilde{\alpha}_{kuv} \|_2 +  \| \tilde{\alpha}_{ku(v+1)} - \tilde{\alpha}_{kuv} \|_2 .
\end{gather}

In summary, text-driven manipulation is conducted by optimizing $P$ and minimizing the following loss
\vspace{-0.05cm}
\begin{equation}
    \mathcal{L}_{edit} = \lambda_{CLIP} \mathcal{L}_{CLIP} + \lambda_{ACR} \mathcal{L}_{ACR} 
\end{equation}

\noindent where $\lambda_{CLIP}$ and $\lambda_{ACR}$ are hyper-parameters.

\begin{figure}
    \includegraphics[width=\linewidth]{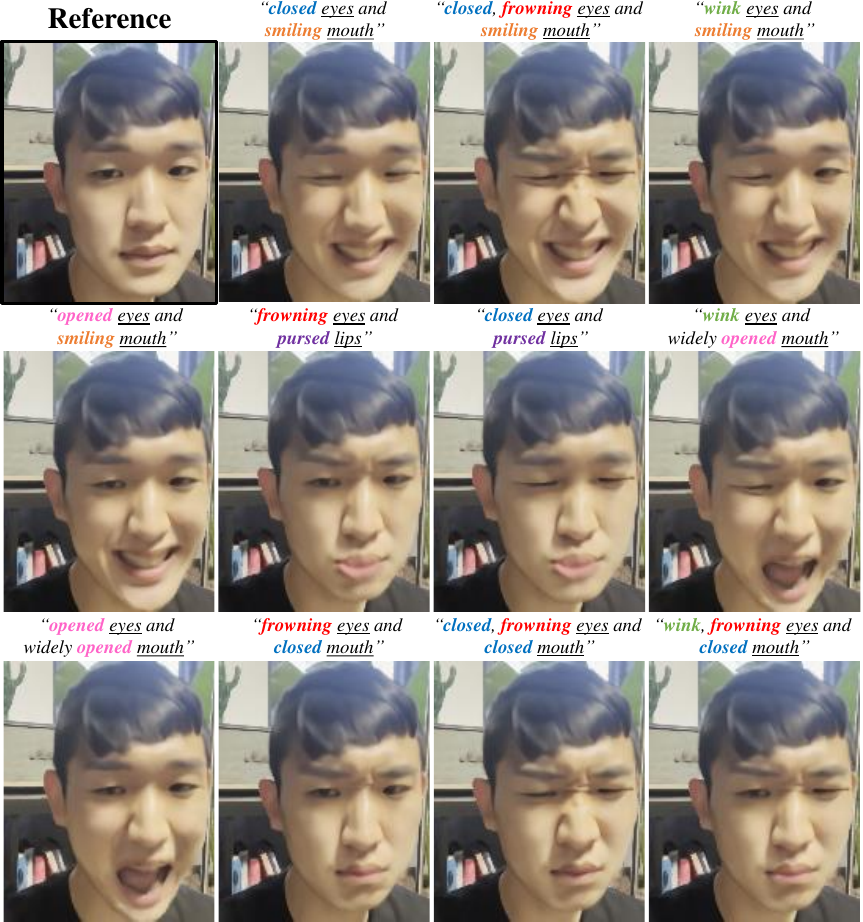}
    \caption{Qualitative results manipulated with descriptive texts using our method. Local facial deformations can easily be controlled using texts only.}
    \label{fig:longtext}
    \vspace{-0.5cm}
\end{figure}

\section{Experiments}

\paragraph{Dataset} We collected portrait videos from six volunteers using Apple iPhone 13, where each volunteer was asked to make four types of facial deformations shown in Fig.~\ref{fig:front}\textcolor{red}{a} and Fig.~\ref{fig:2.b}. A pre-trained human segmentation network was used to exclude descriptors from the dynamic part of the scenes during camera pose computation using COLMAP\cite{schoenberger2016mvs}. Examples of facial deformations observed during training for each scene are reported in the supplementary material.

\begin{figure}
    \includegraphics[width=\linewidth]{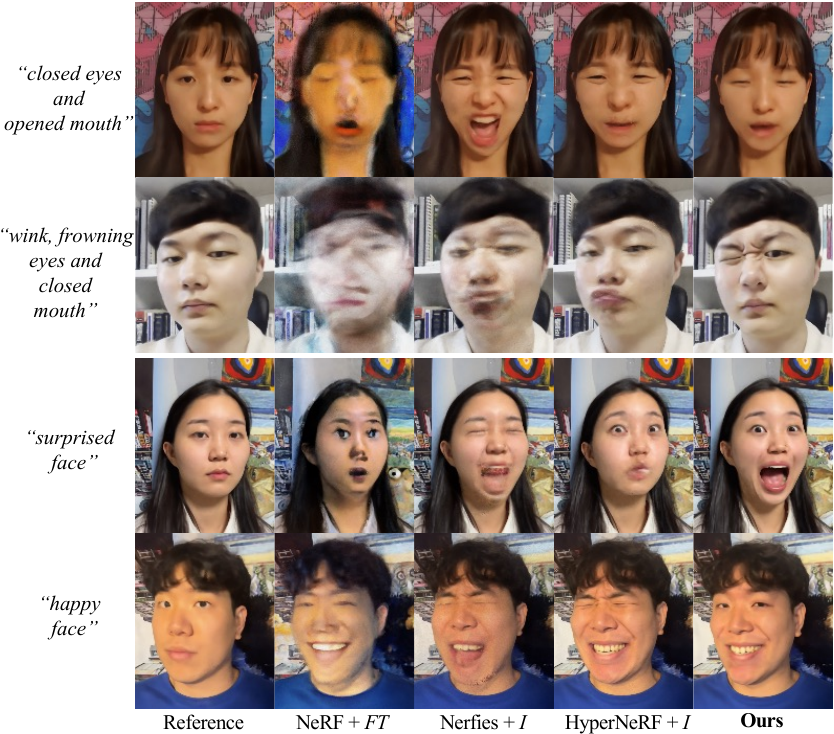}
    \caption{Text-driven manipulation results of our method and the baselines. Our result well reflects the implicit attributes of target emotional texts while preserving visual quality and face identity.}
    \label{fig:all_comparisons}
    \vspace{-0.5cm}
\end{figure}

\vspace{-0.3cm}
\paragraph{Manipulation Texts} We selected two types of texts for manipulation experiments. First is a descriptive text that characterizes deformations of each facial parts. Second is an emotional expression text, which is an implicit representation of a set of multiple local deformations on \textit{all} face parts hard to be described with descriptive texts. We selected 7 frequently used and distinguishable emotional expression texts for our experiment: \textit{"crying", "disappointed", "surprised", "happy", "angry", "scared"} and \textit{"sleeping"}. To reduce text embedding noise, we followed \cite{patashnik2021styleclip} by averaging augmented embeddings of sentences with identical meanings.


\begin{figure*}
  \centering
  \includegraphics[width=\linewidth]{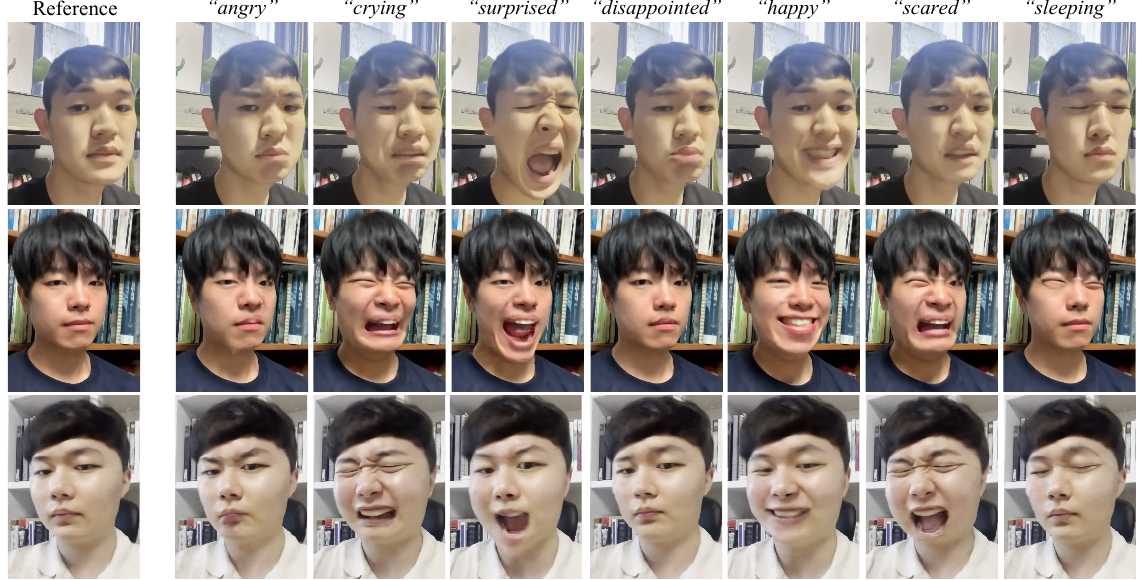}
  \caption{Extensive face manipulation results driven by a set of frequently used emotional expression texts using our method. Manipulating to emotional expression texts are challenging, as they implicitly require compositions of subtle facial deformations that are hard to be described. Our method reasonably reflects the attributes of the manipulation texts.}
  \label{fig:many_results}
  \vspace{-0.3cm}
\end{figure*}

\vspace{-0.3cm}
\paragraph{Baselines} Since there is no prior work that is parallel to our problem definition, we formulated 3 baselines with existing state-of-the-art methods for comparisons: (1) NeRF +\textit{FT} is a simple extension from NeRF \cite{mildenhall2020nerf}  that fine-tunes the whole network using CLIP loss, (2) Nerfies+\textit{I} uses Nerfies\cite{park2021nerfies} as a deformation network followed by conducting vanilla inversion method introduced in Sec. \S\ref{subsec:4.2} for manipulation, and (3) HyperNeRF+\textit{I} replaces Nerfies in (2) with HyperNeRF \cite{park2021hypernerf}.


\vspace{-0.3cm}
\paragraph{Text-driven Manipulation}
We report qualitative manipulation results of our methods driven with a set of descriptive sentences in Fig.~\ref{fig:longtext}. Our method not only faithfully reflects the descriptions, but also can easily control local facial deformations with simple change of words in sentences. We also report manipulated results driven by emotional expression texts in  Fig.~\ref{fig:many_results}.  As can be seen, our method conducts successful manipulations even if the emotional texts are implicit representations of many local facial deformations. For instance, result manipulated with "\textit{crying}" in first row of Fig.~\ref{fig:many_results} is not expressed with mere crying-looking eyes and mouth, but also includes crying-looking eyebrows and skin all over the face without any explicit supervision on local deformations. We also compare our qualitative results to those from the baselines in Fig.~\ref{fig:all_comparisons}. Ours result in the highest reflections of the target text attributes. NeRF+\textit{FT} shows significant degradation in visual quality, while Nerfies+\textit{I} moderately suffers from low reconstruction quality and reflection of target text attributes. HyperNeRF+ \textit{I} shows the highest visual quality out of all baselines, yet fails to reflect the visual attributes of target texts. 

\begin{figure}
    \centering
    \includegraphics[width=0.95\linewidth]{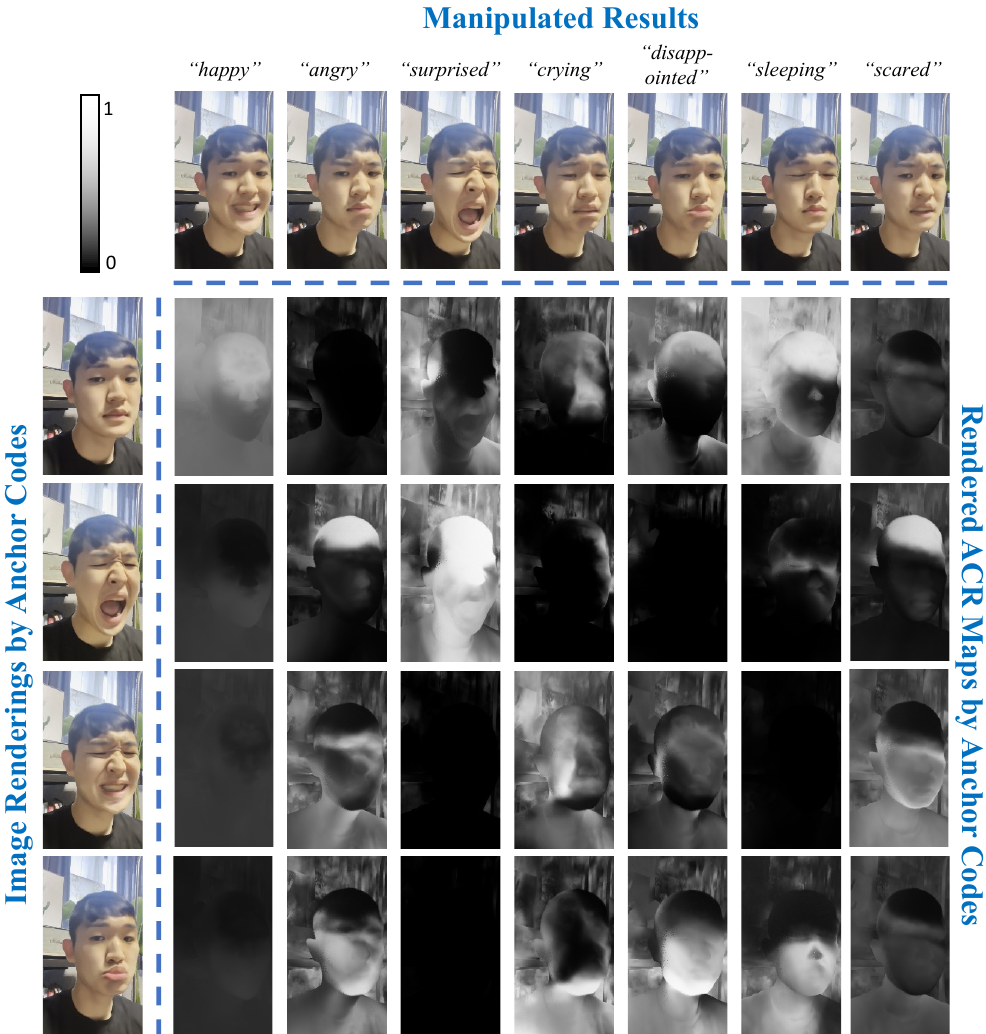}
    \caption{Renderings of learned ACR maps for each anchor codes over different manipulation texts.}
    \label{fig:alpha}
    \vspace{-0.3cm}
\end{figure}

High reflectivity on various manipulation texts can be attributed to PAC that resolves the linked local attribute problem. In Fig.~\ref{fig:alpha}, we visualize $\tilde{\alpha}_{kuv}$ for each anchor code $k$, which is the rendering of ACR $\alpha_{[\textbf{x}, k]}$ in Eq. \eqref{eq:alpha}, over an image plane. Whiter regions of the renderings are closer to one, which indicates that the corresponding anchor code is mostly composited to yield the latent code of the region. Also, we display image renderings from each anchor code on the left to help understand the local attributes for each anchor code. As can be seen, PAC composes appropriate anchor codes for different positions. For example, when manipulating for \textit{sleeping} face, PAC reflects closed eyes from one anchor code and neutral mouth from other anchor codes. In the cases of \textit{crying}, \textit{angry}, \textit{scared}, and \textit{disappointed} face, PAC learns to produce complicated compositions of learned deformations, which are inexpressible with a single latent code.

\begin{table}
  \centering
   \resizebox{\linewidth}{!}{\begin{tabular}{@{}cccc@{}}
    \toprule
    & R-Prec.\cite{xu2018attngan} $\uparrow$ & LPIPS\cite{zhang2018unreasonable} $\downarrow$ & CFS $\uparrow$ \\
    \midrule
    NeRF + \textit{FT} & \underline{\textit{0.763}} & 0.350 & 0.350 \\
    Nerfies + \textit{I} & 0.213 & 0.222 & 0.684 \\
    HyperNeRF + \textit{I} & 0.342 & \underline{\textit{0.198}} & \underline{\textit{0.721}} \\
    \midrule
    Ours & \textbf{0.780} (+0.017)  & \textbf{0.082}  (-0.116) & \textbf{0.749}  (+0.028) \\

    \bottomrule
  \end{tabular}}
  \caption{Quantitative results. R-Prec. denotes R-precision, and CFS denotes cosine face similarity. We notate performance ranks as \textbf{best} and \underline{\textit{second best}}.}
  \label{tab:metrics}

  \resizebox{\linewidth}{!}{\begin{tabular}{@{}cccc@{}}
    \toprule
    & TR $\uparrow$ & VR $\uparrow$ & FP $\uparrow$ \\
    \midrule
    
    NeRF + \textit{FT} & \underline{\textit{2.85}} & 0.18 & 0.79 \\
    Nerfies + \textit{I} & 0.33 & 3.61 & 4.03 \\
    HyperNeRF + \textit{I} & 2.52 & \underline{\textit{4.42}} & \underline{\textit{4.39}} \\
    \midrule
    Ours & \textbf{4.15} (+1.30) & \textbf{4.58} (+0.16) & \textbf{4.67} (+0.28) \\

    \bottomrule
  \end{tabular}}
  \caption{User study results. TR, VR, and FP denote text reflectivity, visual realism, and face identity preservability, respectively. \textbf{Best} and \underline{\textit{second best}} are highlighted.}
  \vspace{-0.4cm}
  \label{tab:user_study}
  
\end{table}

\vspace{-0.3cm}
\paragraph{Quantitative Results} First of all, we measured R-precision\cite{xu2018attngan} to measure the text attribute reflectivity of the manipulations. We used facial expression recognition model\cite{savchenko2022frame} pre-trained with AffectNet\cite{mollahosseini2017affectnet} for top-R retrievals of each text. Specifically, 1000 novel view images are rendered per face, where 200 images are rendered from a face manipulated with each of the five texts that are distinguishable and exist in AffectNet labels: \textit{"happy"}, \textit{"surprised"}, \textit{"fearful"}, \textit{"angry"}, and \textit{"sad"}. Also, to estimate the visual quality after manipulation, we measured LPIPS\cite{zhang2018unreasonable} between faces with no facial expressions (neutral faces) without any manipulations and faces manipulated with 7 texts, each of which are rendered from 200 novel views. Note that LPIPS was our best estimate of visual quality since there can be no pixel-wise ground truth of text-driven manipulations. Lastly, to measure how much of the facial identity is preserved after manipulation, we measured the cosine similarity between face identity features\footnote[1]{https://github.com/ronghuaiyang/arcface-pytorch} extracted from neutral faces and text-manipulated faces, all of which are rendered from 200 novel views. Table \ref{tab:metrics} reports the average results over all texts, which shows that our method outperforms in all criteria. 

\vspace{-0.3cm}
\paragraph{User Study} Users were asked to score from 0 to 5 on 3 criteria; (i) Text Reflectivity: how well the manipulated renderings reflect the target texts, (ii) Visual Realism: how realistic do the manipulated images look, and (iii) Face identity Preservability: how well do the manipulated images preserve the identity of the original face, over our method and the baselines. The following results are reported in Table. \ref{tab:user_study}. Our method outperforms all baselines, and especially in text reflectivity by a large margin. Note that the out-performance in user responses align with that from the quantitative results, which supports the consistency of evaluations.

\begin{figure}
    \centering
    \includegraphics[width=1\linewidth]{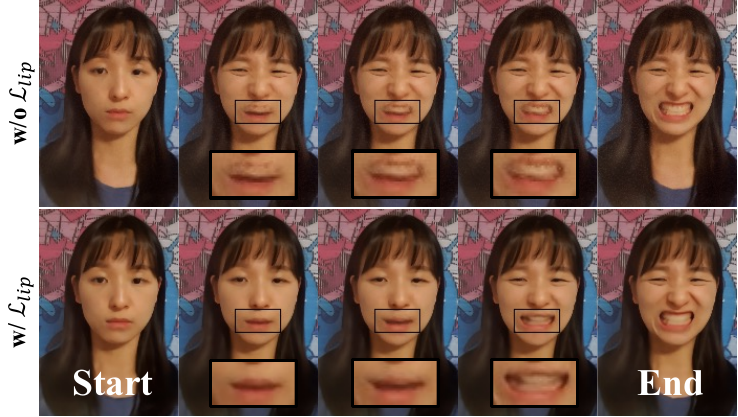}
    \caption{Renderings from linearly interpolated latent codes. Lipschitz-regularized scene manipulator interpolates unseen shapes more naturally.}
    \label{fig:lerp}
    
    \includegraphics[width=\linewidth]{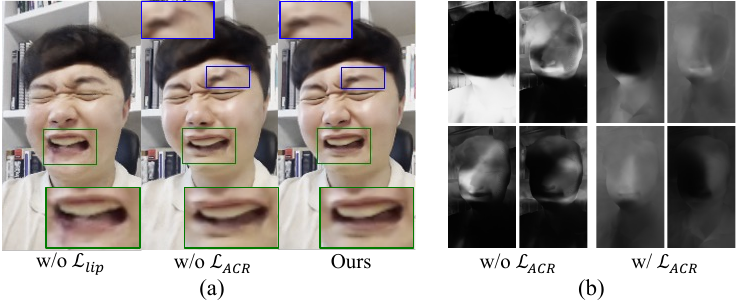}
    \caption{(a) Qualitative results of the ablation study. Manipulations are done using \textit{"crying face"} as target text. (b) Rendered ACR maps with and without $\mathcal{L}_{ACR}$.}
    \label{fig:ablation}
    \vspace{-0.5cm}
\end{figure}

\vspace{-0.3cm}
\paragraph{Interpolation} We experiment with the effect of Lipschitz regularization on the scene manipulator by comparing the visual quality of images rendered from linearly interpolated latent codes, and report the results in Fig.~\ref{fig:lerp}. Lipschitz-regularized scene manipulator yields more visually natural images, which implies that learned set of anchor-composited latent codes $[w^{*}_\textbf{x}]$ are more likely to render realistically interpolated local deformations under Lipschitz-regularized scene manipulator. 

\vspace{-0.3cm}
\paragraph{Ablation Study} We conducted an ablation study on our regularization methods: $\mathcal{L}_{lip}$ and $\mathcal{L}_{ACR}$. As shown in Fig.~\ref{fig:ablation}\textcolor{red}{a}, manipulation without $\mathcal{L}_{lip}$ suffers from low visual quality. Manipulation without $\mathcal{L}_{ACR}$ yields unnatural renderings of face parts with large deformation range such as mouth and eyebrows. This can be interpreted with learned ACR maps of PAC in Fig.~\ref{fig:ablation}\textcolor{red}{b}. ACR maps learned with $\mathcal{L}_{ACR}$ introduces reasonable continuities of latent codes on boundaries of the dynamic face parts, thus yielding naturally interpolated face parts.  


\vspace{-0.15cm}
\section{Conclusion}
We have presented FaceCLIPNeRF, a text-driven manipulation pipeline of a 3D face using deformable NeRF. We first proposed a Lipshitz-regularized scene manipulator, a conditional MLP that uses its latent code as a control handle of facial deformations. We addressed the linked local attribute problem of conventional deformable NeRFs, which cannot compose deformations observed in different instances. As so, we proposed PAC that learns to produce spatially-varying latent codes, whose renderings with the scene manipulator were trained to yield high cosine similarity with target text in CLIP embedding space. Our experiments showed that our method could faithfully reflect the visual attributes of both descriptive and emotional texts while preserving visual quality and identity of 3D face. 

\paragraph{Acknowledgement} This material is based upon work supported by the Air Force Office of Scientific Research under award number FA2386-22-1-4024, KAIST-NAVER hypercreative AI center, and the Institute of Information \& communications Technology Planning \& Evaluation (IITP) grant funded by the Korea government (MSIT) (No.2019-0-00075, Artificial Intelligence Graduate School Program (KAIST)).





{\small
\bibliographystyle{ieee_fullname}
\bibliography{egbib}

\begin{thebibliography}{10}\itemsep=-1pt

\bibitem{barron2021mip}
Jonathan~T Barron, Ben Mildenhall, Matthew Tancik, Peter Hedman, Ricardo
  Martin-Brualla, and Pratul~P Srinivasan.
\newblock Mip-nerf: A multiscale representation for anti-aliasing neural
  radiance fields.
\newblock In {\em Proceedings of the IEEE/CVF International Conference on
  Computer Vision}, pages 5855--5864, 2021.

\bibitem{blanz1999morphable}
Volker Blanz and Thomas Vetter.
\newblock A morphable model for the synthesis of 3d faces.
\newblock In {\em Proceedings of the 26th annual conference on Computer
  graphics and interactive techniques}, pages 187--194, 1999.

\bibitem{ester1996density}
Martin Ester, Hans-Peter Kriegel, J{\"o}rg Sander, Xiaowei Xu, et~al.
\newblock A density-based algorithm for discovering clusters in large spatial
  databases with noise.
\newblock In {\em kdd}, volume~96, pages 226--231, 1996.

\bibitem{fan2022unified}
Zhiwen Fan, Yifan Jiang, Peihao Wang, Xinyu Gong, Dejia Xu, and Zhangyang Wang.
\newblock Unified implicit neural stylization.
\newblock {\em arXiv preprint arXiv:2204.01943}, 2022.

\bibitem{DECA:Siggraph2021}
Yao Feng, Haiwen Feng, Michael~J. Black, and Timo Bolkart.
\newblock Learning an animatable detailed {3D} face model from in-the-wild
  images.
\newblock volume~40, 2021.

\bibitem{gafni2021dynamic}
Guy Gafni, Justus Thies, Michael Zollhofer, and Matthias Nie{\ss}ner.
\newblock Dynamic neural radiance fields for monocular 4d facial avatar
  reconstruction.
\newblock In {\em Proceedings of the IEEE/CVF Conference on Computer Vision and
  Pattern Recognition}, pages 8649--8658, 2021.

\bibitem{hong2022headnerf}
Yang Hong, Bo Peng, Haiyao Xiao, Ligang Liu, and Juyong Zhang.
\newblock Headnerf: A real-time nerf-based parametric head model.
\newblock In {\em Proceedings of the IEEE/CVF Conference on Computer Vision and
  Pattern Recognition}, pages 20374--20384, 2022.

\bibitem{hormann2014barycentric}
Kai Hormann.
\newblock Barycentric interpolation.
\newblock In {\em Approximation Theory XIV: San Antonio 2013}, pages 197--218.
  Springer, 2014.

\bibitem{jain2022zero}
Ajay Jain, Ben Mildenhall, Jonathan~T Barron, Pieter Abbeel, and Ben Poole.
\newblock Zero-shot text-guided object generation with dream fields.
\newblock In {\em Proceedings of the IEEE/CVF Conference on Computer Vision and
  Pattern Recognition}, pages 867--876, 2022.

\bibitem{jain2021dreamfields}
Ajay Jain, Ben Mildenhall, Jonathan~T. Barron, Pieter Abbeel, and Ben Poole.
\newblock Zero-shot text-guided object generation with dream fields.
\newblock 2022.

\bibitem{jetchev2021clipmatrix}
Nikolay Jetchev.
\newblock Clipmatrix: Text-controlled creation of 3d textured meshes.
\newblock {\em arXiv preprint arXiv:2109.12922}, 2021.

\bibitem{kajiya1984ray}
James~T Kajiya and Brian~P Von~Herzen.
\newblock Ray tracing volume densities.
\newblock {\em ACM SIGGRAPH computer graphics}, 18(3):165--174, 1984.

\bibitem{kania2022conerf}
Kacper Kania, Kwang~Moo Yi, Marek Kowalski, Tomasz Trzci{\'n}ski, and Andrea
  Tagliasacchi.
\newblock {CoNeRF: Controllable Neural Radiance Fields}.
\newblock In {\em Proceedings of the IEEE Conference on Computer Vision and
  Pattern Recognition}, 2022.

\bibitem{kobayashi2022distilledfeaturefields}
Sosuke Kobayashi, Eiichi Matsumoto, and Vincent Sitzmann.
\newblock Decomposing nerf for editing via feature field distillation.
\newblock In {\em Advances in Neural Information Processing Systems},
  volume~35, 2022.

\bibitem{lazova2022control}
Verica Lazova, Vladimir Guzov, Kyle Olszewski, Sergey Tulyakov, and Gerard
  Pons-Moll.
\newblock Control-nerf: Editable feature volumes for scene rendering and
  manipulation.
\newblock {\em arXiv preprint arXiv:2204.10850}, 2022.

\bibitem{li2022neural}
Tianye Li, Mira Slavcheva, Michael Zollhoefer, Simon Green, Christoph Lassner,
  Changil Kim, Tanner Schmidt, Steven Lovegrove, Michael Goesele, Richard
  Newcombe, et~al.
\newblock Neural 3d video synthesis from multi-view video.
\newblock In {\em Proceedings of the IEEE/CVF Conference on Computer Vision and
  Pattern Recognition}, pages 5521--5531, 2022.

\bibitem{liu2022learning}
Hsueh-Ti~Derek Liu, Francis Williams, Alec Jacobson, Sanja Fidler, and Or
  Litany.
\newblock Learning smooth neural functions via lipschitz regularization.
\newblock {\em arXiv preprint arXiv:2202.08345}, 2022.

\bibitem{liu2021editing}
Steven Liu, Xiuming Zhang, Zhoutong Zhang, Richard Zhang, Jun-Yan Zhu, and
  Bryan Russell.
\newblock Editing conditional radiance fields.
\newblock In {\em Proceedings of the IEEE/CVF International Conference on
  Computer Vision}, pages 5773--5783, 2021.

\bibitem{mildenhall2022nerf}
Ben Mildenhall, Peter Hedman, Ricardo Martin-Brualla, Pratul~P Srinivasan, and
  Jonathan~T Barron.
\newblock Nerf in the dark: High dynamic range view synthesis from noisy raw
  images.
\newblock In {\em Proceedings of the IEEE/CVF Conference on Computer Vision and
  Pattern Recognition}, pages 16190--16199, 2022.

\bibitem{mildenhall2020nerf}
Ben Mildenhall, Pratul~P. Srinivasan, Matthew Tancik, Jonathan~T. Barron, Ravi
  Ramamoorthi, and Ren Ng.
\newblock Nerf: Representing scenes as neural radiance fields for view
  synthesis.
\newblock In {\em ECCV}, 2020.

\bibitem{mollahosseini2017affectnet}
Ali Mollahosseini, Behzad Hasani, and Mohammad~H Mahoor.
\newblock Affectnet: A database for facial expression, valence, and arousal
  computing in the wild.
\newblock {\em IEEE Transactions on Affective Computing}, 10(1):18--31, 2017.

\bibitem{park2021nerfies}
Keunhong Park, Utkarsh Sinha, Jonathan~T Barron, Sofien Bouaziz, Dan~B Goldman,
  Steven~M Seitz, and Ricardo Martin-Brualla.
\newblock Nerfies: Deformable neural radiance fields.
\newblock In {\em Proceedings of the IEEE/CVF International Conference on
  Computer Vision}, pages 5865--5874, 2021.

\bibitem{park2021hypernerf}
Keunhong Park, Utkarsh Sinha, Peter Hedman, Jonathan~T. Barron, Sofien Bouaziz,
  Dan~B Goldman, Ricardo Martin-Brualla, and Steven~M. Seitz.
\newblock Hypernerf: A higher-dimensional representation for topologically
  varying neural radiance fields.
\newblock {\em ACM Trans. Graph.}, 40(6), dec 2021.

\bibitem{patashnik2021styleclip}
Or Patashnik, Zongze Wu, Eli Shechtman, Daniel Cohen-Or, and Dani Lischinski.
\newblock Styleclip: Text-driven manipulation of stylegan imagery.
\newblock In {\em International Conference of Computer Vision}, pages
  2085--2094, 2021.

\bibitem{poole2022dreamfusion}
Ben Poole, Ajay Jain, Jonathan~T Barron, and Ben Mildenhall.
\newblock Dreamfusion: Text-to-3d using 2d diffusion.
\newblock {\em arXiv preprint arXiv:2209.14988}, 2022.

\bibitem{pumarola2020d}
Albert Pumarola, Enric Corona, Gerard Pons-Moll, and Francesc Moreno-Noguer.
\newblock D-nerf: Neural radiance fields for dynamic scenes.
\newblock {\em arXiv preprint arXiv:2011.13961}, 2020.

\bibitem{radford2021learning}
Alec Radford, Jong~Wook Kim, Chris Hallacy, Aditya Ramesh, Gabriel Goh,
  Sandhini Agarwal, Girish Sastry, Amanda Askell, Pamela Mishkin, Jack Clark,
  et~al.
\newblock Learning transferable visual models from natural language
  supervision.
\newblock In {\em International Conference on Machine Learning}, pages
  8748--8763. PMLR, 2021.

\bibitem{ramon2021h3d}
Eduard Ramon, Gil Triginer, Janna Escur, Albert Pumarola, Jaime Garcia, Xavier
  Giro-i Nieto, and Francesc Moreno-Noguer.
\newblock H3d-net: Few-shot high-fidelity 3d head reconstruction.
\newblock In {\em Proceedings of the IEEE/CVF International Conference on
  Computer Vision}, pages 5620--5629, 2021.

\bibitem{sanghi2022clip}
Aditya Sanghi, Hang Chu, Joseph~G Lambourne, Ye Wang, Chin-Yi Cheng, Marco
  Fumero, and Kamal~Rahimi Malekshan.
\newblock Clip-forge: Towards zero-shot text-to-shape generation.
\newblock In {\em Proceedings of the IEEE/CVF Conference on Computer Vision and
  Pattern Recognition}, pages 18603--18613, 2022.

\bibitem{savchenko2022frame}
Andrey~V Savchenko.
\newblock Frame-level prediction of facial expressions, valence, arousal and
  action units for mobile devices.
\newblock {\em arXiv preprint arXiv:2203.13436}, 2022.

\bibitem{schoenberger2016mvs}
Johannes~Lutz Sch\"{o}nberger, Enliang Zheng, Marc Pollefeys, and Jan-Michael
  Frahm.
\newblock {Pixelwise View Selection for Unstructured Multi-View Stereo}.
\newblock In {\em European Conference on Computer Vision (ECCV)}, 2016.

\bibitem{sharma20223d}
Sahil Sharma and Vijay Kumar.
\newblock 3d face reconstruction in deep learning era: A survey.
\newblock {\em Archives of Computational Methods in Engineering}, pages 1--33,
  2022.

\bibitem{sun2022ide}
Jingxiang Sun, Xuan Wang, Yichun Shi, Lizhen Wang, Jue Wang, and Yebin Liu.
\newblock Ide-3d: Interactive disentangled editing for high-resolution 3d-aware
  portrait synthesis.
\newblock {\em arXiv preprint arXiv:2205.15517}, 2022.

\bibitem{sun2022fenerf}
Jingxiang Sun, Xuan Wang, Yong Zhang, Xiaoyu Li, Qi Zhang, Yebin Liu, and Jue
  Wang.
\newblock Fenerf: Face editing in neural radiance fields.
\newblock In {\em Proceedings of the IEEE/CVF Conference on Computer Vision and
  Pattern Recognition}, pages 7672--7682, 2022.

\bibitem{tancik2021learned}
Matthew Tancik, Ben Mildenhall, Terrance Wang, Divi Schmidt, Pratul~P
  Srinivasan, Jonathan~T Barron, and Ren Ng.
\newblock Learned initializations for optimizing coordinate-based neural
  representations.
\newblock In {\em Proceedings of the IEEE/CVF Conference on Computer Vision and
  Pattern Recognition}, pages 2846--2855, 2021.

\bibitem{thies2016face2face}
Justus Thies, Michael Zollhofer, Marc Stamminger, Christian Theobalt, and
  Matthias Nie{\ss}ner.
\newblock Face2face: Real-time face capture and reenactment of rgb videos.
\newblock In {\em Proceedings of the IEEE conference on computer vision and
  pattern recognition}, pages 2387--2395, 2016.

\bibitem{verbin2022refnerf}
Dor Verbin, Peter Hedman, Ben Mildenhall, Todd Zickler, Jonathan~T. Barron, and
  Pratul~P. Srinivasan.
\newblock {Ref-NeRF}: Structured view-dependent appearance for neural radiance
  fields.
\newblock {\em CVPR}, 2022.

\bibitem{wang2022clip}
Can Wang, Menglei Chai, Mingming He, Dongdong Chen, and Jing Liao.
\newblock Clip-nerf: Text-and-image driven manipulation of neural radiance
  fields.
\newblock In {\em Proceedings of the IEEE/CVF Conference on Computer Vision and
  Pattern Recognition}, pages 3835--3844, 2022.

\bibitem{wang2021learning}
Ziyan Wang, Timur Bagautdinov, Stephen Lombardi, Tomas Simon, Jason Saragih,
  Jessica Hodgins, and Michael Zollhofer.
\newblock Learning compositional radiance fields of dynamic human heads.
\newblock In {\em Proceedings of the IEEE/CVF Conference on Computer Vision and
  Pattern Recognition}, pages 5704--5713, 2021.

\bibitem{xu2018attngan}
Tao Xu, Pengchuan Zhang, Qiuyuan Huang, Han Zhang, Zhe Gan, Xiaolei Huang, and
  Xiaodong He.
\newblock Attngan: Fine-grained text to image generation with attentional
  generative adversarial networks.
\newblock In {\em Proceedings of the IEEE conference on computer vision and
  pattern recognition}, pages 1316--1324, 2018.

\bibitem{yang2021objectnerf}
Bangbang Yang, Yinda Zhang, Yinghao Xu, Yijin Li, Han Zhou, Hujun Bao, Guofeng
  Zhang, and Zhaopeng Cui.
\newblock Learning object-compositional neural radiance field for editable
  scene rendering.
\newblock In {\em International Conference on Computer Vision ({ICCV})},
  October 2021.

\bibitem{yenamandra2021i3dmm}
Tarun Yenamandra, Ayush Tewari, Florian Bernard, Hans-Peter Seidel, Mohamed
  Elgharib, Daniel Cremers, and Christian Theobalt.
\newblock i3dmm: Deep implicit 3d morphable model of human heads.
\newblock In {\em Proceedings of the IEEE/CVF Conference on Computer Vision and
  Pattern Recognition}, pages 12803--12813, 2021.

\bibitem{yoshida2017spectral}
Yuichi Yoshida and Takeru Miyato.
\newblock Spectral norm regularization for improving the generalizability of
  deep learning.
\newblock {\em arXiv preprint arXiv:1705.10941}, 2017.

\bibitem{yu2021plenoctrees}
Alex Yu, Ruilong Li, Matthew Tancik, Hao Li, Ren Ng, and Angjoo Kanazawa.
\newblock Plenoctrees for real-time rendering of neural radiance fields.
\newblock In {\em Proceedings of the IEEE/CVF International Conference on
  Computer Vision}, pages 5752--5761, 2021.

\bibitem{yuan2022nerf}
Yu-Jie Yuan, Yang-Tian Sun, Yu-Kun Lai, Yuewen Ma, Rongfei Jia, and Lin Gao.
\newblock Nerf-editing: geometry editing of neural radiance fields.
\newblock In {\em Proceedings of the IEEE/CVF Conference on Computer Vision and
  Pattern Recognition}, pages 18353--18364, 2022.

\bibitem{zhang2018unreasonable}
Richard Zhang, Phillip Isola, Alexei~A Efros, Eli Shechtman, and Oliver Wang.
\newblock The unreasonable effectiveness of deep features as a perceptual
  metric.
\newblock In {\em Proceedings of the IEEE conference on computer vision and
  pattern recognition}, pages 586--595, 2018.

\bibitem{zheng2022avatar}
Yufeng Zheng, Victoria~Fern{\'a}ndez Abrevaya, Marcel~C B{\"u}hler, Xu Chen,
  Michael~J Black, and Otmar Hilliges.
\newblock Im avatar: Implicit morphable head avatars from videos.
\newblock In {\em Proceedings of the IEEE/CVF Conference on Computer Vision and
  Pattern Recognition}, pages 13545--13555, 2022.

\bibitem{zhuang2021mofanerf}
Yiyu Zhuang, Hao Zhu, Xusen Sun, and Xun Cao.
\newblock Mofanerf: Morphable facial neural radiance field.
\newblock {\em arXiv preprint arXiv:2112.02308}, 2021.

\end{thebibliography}
}

\end{document}


\title{FaceCLIPNeRF: Text-driven 3D Face Manipulation \\ using Deformable Neural Radiance Fields - Supplements}

\author{Sungwon Hwang\textsuperscript{1}\qquad Junha Hyung\textsuperscript{1}\qquad Min-Jung Kim\textsuperscript{1}\qquad Daejin Kim\textsuperscript{1}\qquad Jaegul Choo\textsuperscript{1}\\ 
Graduate School of AI, KAIST\textsuperscript{1}\\
{\tt\small \{shwang.14, sharpeeee, emjay73, kiddj, jchoo\}@kaist.ac.kr}
}

\twocolumn[{
\maketitle
\begin{center}
  \centering
  \captionsetup{type=figure}
  \includegraphics[width=\linewidth]{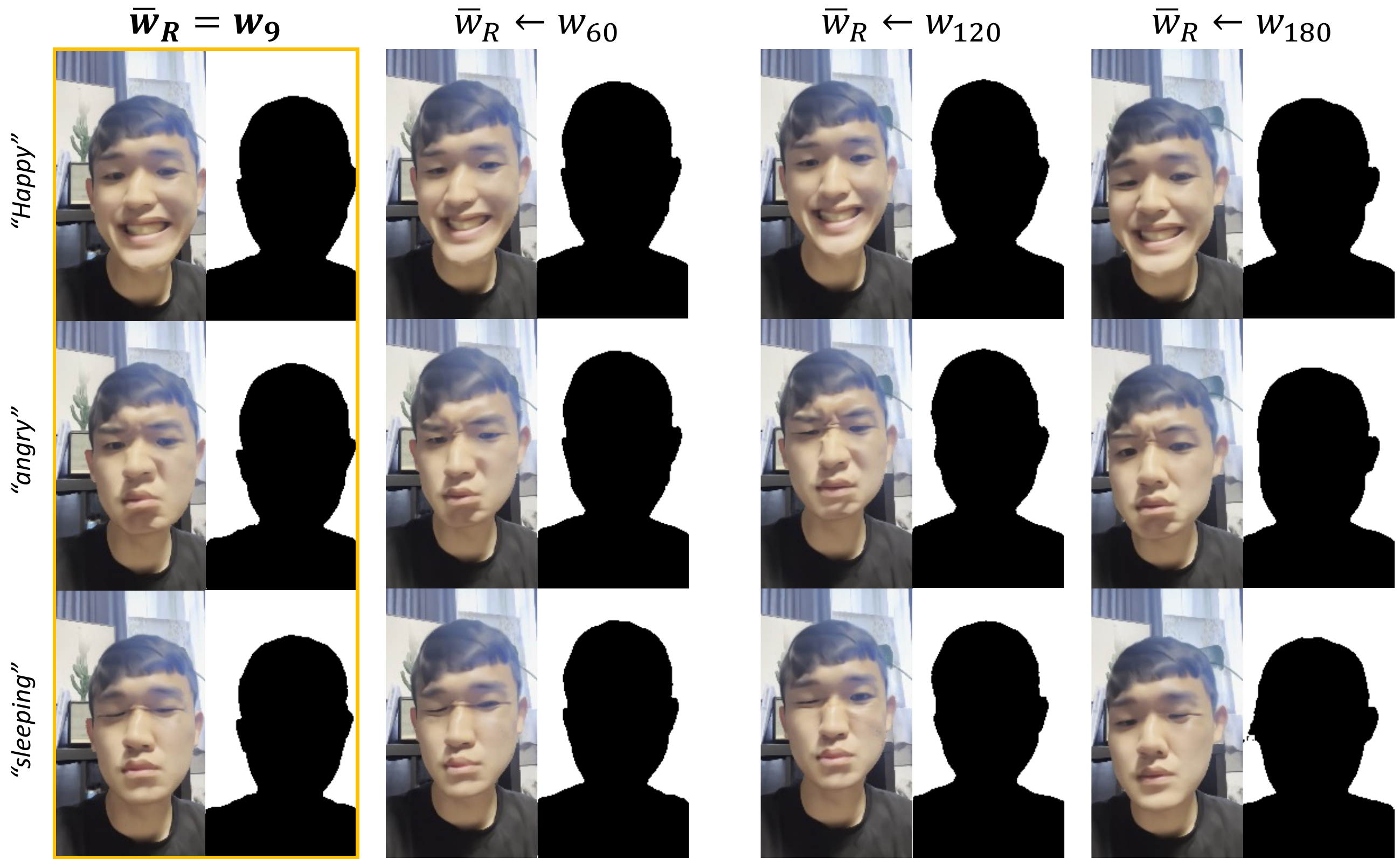}
  \captionof{figure}{Manipulations rendered with different $\bar{w}_{R}$ for deformation field $T$. Our manipulation pipelines for all texts were trained with learned latent code of $9$-th training frame, $\bar{w}_{9}$. Then, we replaced $\bar{w}_{9}$ with $\bar{w}_{60}$, $\bar{w}_{120}$ and $\bar{w}_{180}$ to observe any difference in renderings. Human segmentation masks are also provided for clearer visualization of head poses.}
  \label{fig:sup2}
\end{center}
}]


\section{Constant Latent Code for Deformation Field}

We learned from experiments that fixing $\bar{w}_{R}$, a latent code to deformation field $T$, creates strong controllability of head pose. For example, we conducted text-driven manipulation tasks by allocating fixed $\bar{w}_{9}$, a trained latent code of 9-th training frame, to $\bar{w}_{R}$. After training PAC for manipulations, we replaced $\bar{w}_{R}$ with $\bar{w}_{60}$, $\bar{w}_{120}$, and $\bar{w}_{180}$ during inference over the manipulated scenes to observe any difference. Following results are shown in Fig. \ref{fig:sup2}, where we provide human segmentation masks for clearer visualizations of head poses. One could observe that head poses are relatively constant given a constant $\bar{w}_{R}$ regardless of the text used for our manipulation method, whereas head poses do change when using different latent code for $\bar{w}_{R}$. We made a qualitative measure to verify such observations by measuring two metrics: Intra-mIoU and Inter-mIoU. Intra-mIoU is measured by calculating mIoU between all possible pairs of human segmentation masks extracted from the renderings of the same latent code for $\bar{w}_{R}$. We calculate the average on manipulation networks trained with all 7 proposed text prompts, and the images used to extract human segmentation mask are all rendered in 200 novel views. Meanwhile, Inter-mIoU is calculated using pairs of segmentation masks extracted from different latent codes for $\bar{w}_{R}$. Following results are reported in Table. \ref{tab:iou}. Intra-mIoU is close to $1$, meaning that the head poses are almost constant for constant $\bar{w}_{R}$, while Inter-mIoU is smaller than Intra-mIoU by large amount ($0.170$), which means that head poses change more drastically when using different codes for $\bar{w}_{R}$. 

Another observation noticed in Fig. \ref{fig:sup2} is that facial expressions slightly change when replacing different latent code for $\bar{w}_{R}$, even when there is no change in PAC network and anchor codes. We may conclude from the observation that $T$ intervenes over both rigid deformations such as head pose and detailed deformations such as facial expressions, whereas $H$ only controls detailed facial expressions. We may leave the problem of selecting optimal latent code for $\bar{w}_{R}$ or explicit head control fully disentangled from facial expression for future research topics. 

\begin{table}
  \centering
  \begin{tabular}{@{}cc@{}}
    \toprule
    Intra-mIoU & Inter-mIoU \\
    \midrule
    0.986 & 0.816 \\
    \bottomrule
  \end{tabular}
  \caption{Intra-mIoU and Inter-mIoU calculated with human segmentation masks extracted from images rendered with 4 different latent codes for $\bar{w}_{R}$ fed to deformation network $T$. Higher Intra-mIoU means that more constant head poses are rendered over the same $\bar{w}_{R}$, whereas lower Inter-mIoU means that head poses changes more for different $\bar{w}_{R}$.}
  \label{tab:iou}
\end{table}

\section{Observed facial deformations in our Dataset}

\begin{figure}
  \centering
  \includegraphics[width=\linewidth]{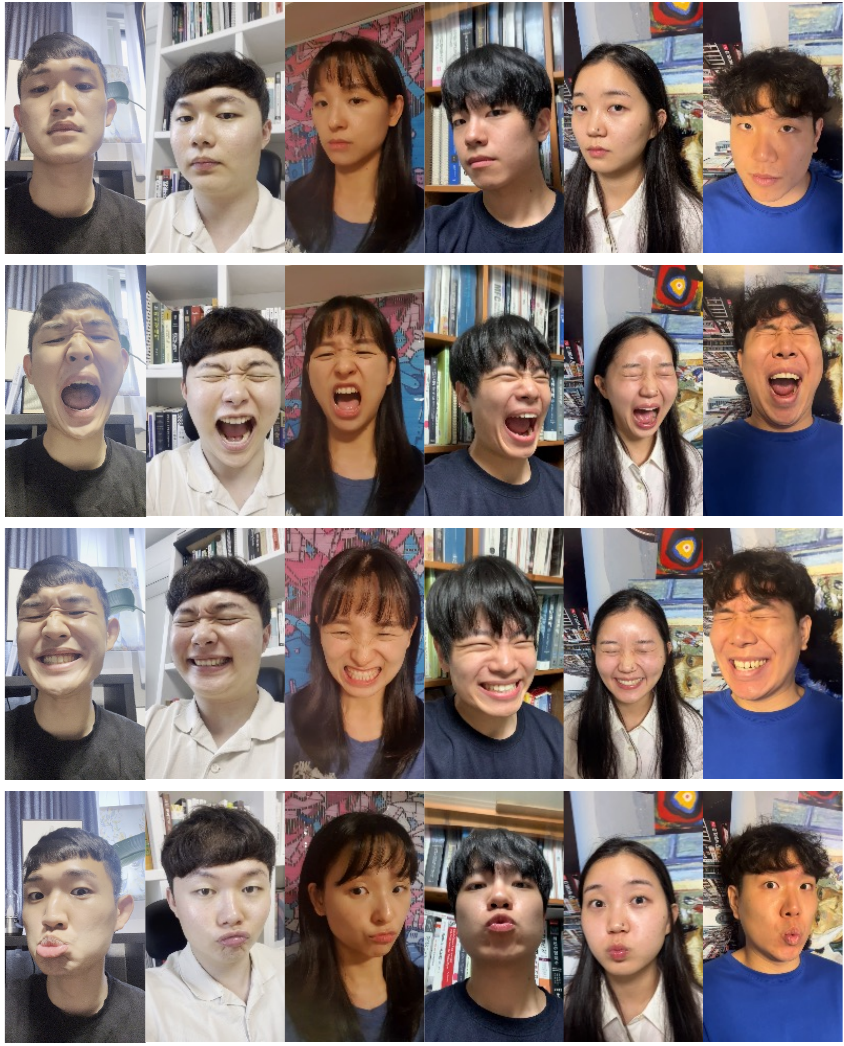}

  \caption{Observed facial deformations in our dataset used for the experiments. All volunteers were asked to make the following facial expressions during the data collection phases.}
  \label{fig:sup1}
\end{figure}

In this section, we report the types of facial deformations observed during the scene manipulator training in Fig. \ref{fig:sup1}. We show that only 4 types of facial deformations are set to be available for each scene. As one of the key motivations of our work is to make the manipulation user-friendly, requiring the subjects to make only a few types of deformation and to collect approximately 300 frames (10 seconds video for 30fps) greatly reduces the amount of labor for non-expert users during data collection phase. 

However, a recommended requirement of the observed facial deformations of a scene is that each face part is provided with at least two oppositely extreme deformations, so that locally composited latent codes can express wider range of interpolated, thus unseen, types of local deformations. For example, a data is required to include at least closed eyes and widely opened eyes to express as many types of eye deformations in between. Same applies to other face parts such as mouth, eyebrows and skin between the eyebrows. We learned that the 4 types of the selected facial deformations are enough to cover the wide range of deformations for each local face part, yet further researches can be done to find optimal set of facial deformation types.

\begin{table*}[h]
  \centering
  \begin{tabular}{@{}cccccc@{}}
    \toprule
    & \textit{"happy"} & \textit{"surprised"} & \textit{"fearful"} & \textit{"angry"} & \textit{"crying"} \\
    \midrule
    NeRF + \textit{FT} & \underline{\textit{0.908}} & \underline{\textit{0.809}} & \textbf{0.503} & \underline{\textit{0.768}} & \underline{\textit{0.828}} \\
    Nerfies + \textit{I} & 0.235 & 0.166 & 0.205 & 0.193 & 0.264 \\
    HyperNeRF + \textit{I} & 0.771 & 0.263 & 0.154 & 0.233 & 0.288 \\
    \midrule
    Ours & \textbf{1.000} & \textbf{0.843} & \underline{\textit{0.313}} & \textbf{0.849} & \textbf{0.898}\\

    \bottomrule
  \end{tabular}
  \caption{R-Precision\cite{xu2018attngan} by different text prompts for manipulation}
  \label{tab:sup_rprec}
  \vspace{0.4cm}
  \begin{tabular}{@{}cccccccc@{}}
    \toprule
    & \textit{"happy"} & \textit{"disappointed"} &\textit{"surprised"} & \textit{"scared"} & \textit{"angry"} & \textit{"crying"} & \textit{"sleeping"} \\
    \midrule
    NeRF + \textit{FT} & 0.337 & 0.349 & 0.342 & 0.356 & 0.351 & 0.342 & 0.376 \\
    Nerfies + \textit{I} & \underline{\textit{0.192}} & 0.218 & \underline{\textit{0.188}} & 0.246 & 0.243 & 0.232 & 0.234 \\
    HyperNeRF + \textit{I} & 0.222 & \underline{\textit{0.171}} & 0.192 & \underline{\textit{0.206}} & \underline{\textit{0.196}} & \underline{\textit{0.208}}& \underline{\textit{0.191}} \\
    \midrule
    Ours & \textbf{0.090} & \textbf{0.052 }& \textbf{0.107} & \textbf{0.089} & \textbf{0.070} & \textbf{0.097} & \textbf{0.070}\\

    \bottomrule
  \end{tabular}
  \caption{LPIPS\cite{zhang2018unreasonable} by different text prompts for manipulation}
  \label{tab:sup_lpips}
  \vspace{0.4cm}
  \begin{tabular}{@{}cccccccc@{}}
    \toprule
    & \textit{"happy"} & \textit{"disappointed"} &\textit{"surprised"} & \textit{"scared"} & \textit{"angry"} & \textit{"crying"} & \textit{"sleeping"} \\
    \midrule
    NeRF + \textit{FT} & 0.466 &0.306 &0.361 &0.296 &0.246 &0.364 &0.410 \\
    Nerfies + \textit{I} & \underline{\textit{0.696}} & 0.722 & 0.655 & 0.678 & \underline{\textit{0.668}} & \textbf{0.690} & 0.676 \\
    HyperNeRF + \textit{I} & 0.672 & \textbf{0.859} & \underline{\textit{0.690}} & \underline{\textit{0.695}} & 0.650 & 0.685 & \underline{\textit{0.793}}\\
    \midrule
    Ours &\textbf{0.785}& \underline{\textit{0.837}} & \textbf{0.712} & \textbf{0.704} &\textbf{ 0.787} & \underline{\textit{0.613}} & \textbf{0.806}\\

    \bottomrule
  \end{tabular}
  \caption{Cosine Face Similarity by different text prompts for manipulation}
  \label{tab:sup_cfs}

\end{table*}

\section{Quantitative Results by Texts}

Here, we report qualitative metrics on each text prompt selected for text-driven manipulation tasks. For R-Precision \cite{xu2018attngan}, our approach outperformed baseline methods except for the text \textit{"fearful"}. However, considering that NeRF + \textit{FT} fine-tunes the whole network to approach its renderings to target text in CLIP embedding, it has more capacity to over-fit to text attribute reflection at the cost of visual quality and face identity preservation. In addition, our approach out-performed all baselines in LPIPS\cite{zhang2018unreasonable}, which implies the contribution of Lipschitz regularization and alpha total variation loss for high visual quality of rendered images. Ours outperformed in Cosine Face Similarity in most text prompts. However, ours showed a relatively low performance in \textit{"crying"}. Such result can be attributed to the ability of our method to render unseen facial deformations by compositing local face deformations observed in different instances. Since the word \textit{"crying"} requires complicated compositions of local deformations during manipulation, the face identity may have varied slightly. However, not only the difference is minute, but ours outperformed in most text prompts in all metrics. As so, we may conclude that our approach is the best option for a text-driven manipulation task of a face in NeRF.

\section{More Qualitative Results and Comparisons to Baselines}

\begin{figure*}
  \centering
  \includegraphics[width=\linewidth]{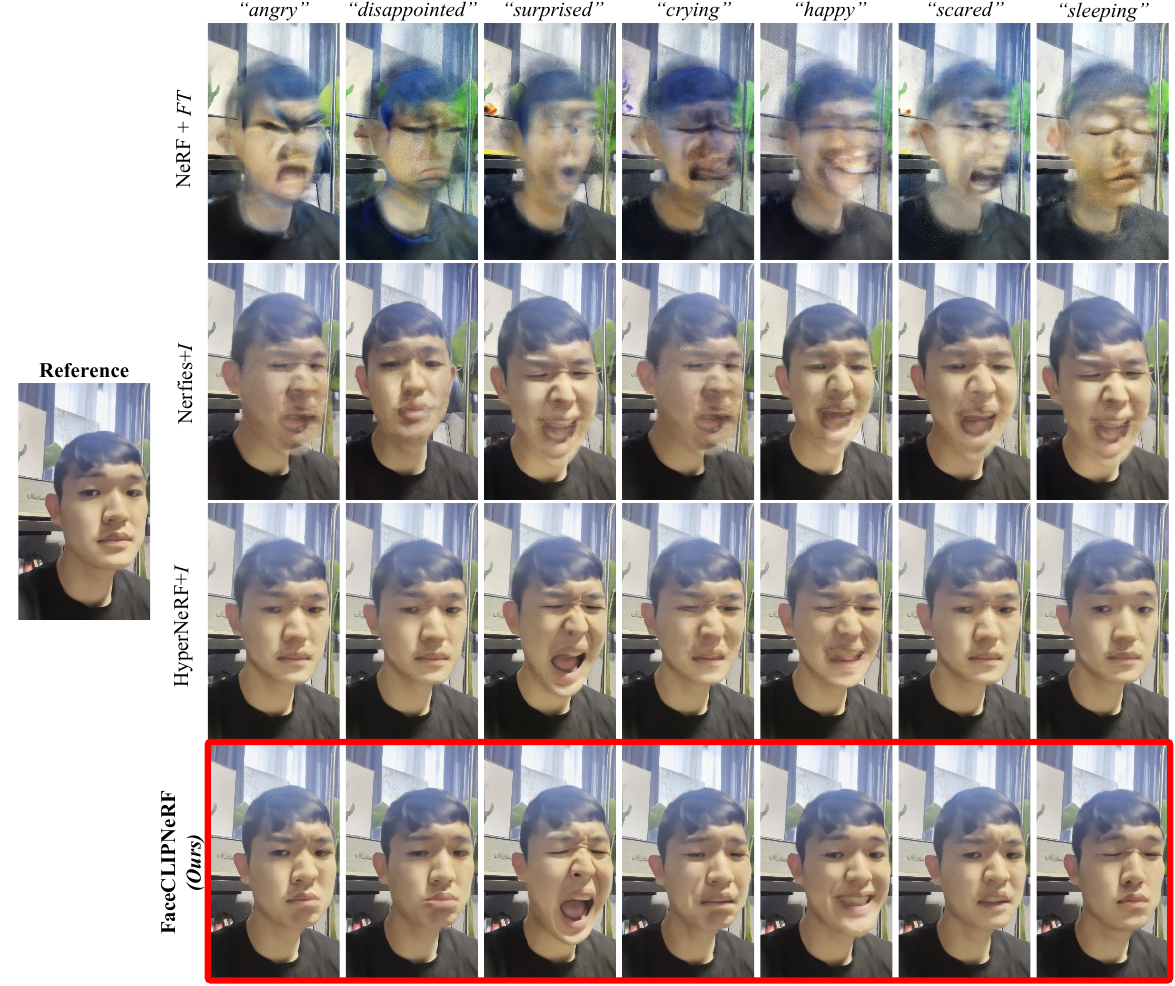}
  \caption{Text-driven manipulation results on volunteer \#1 using baseline methods and our approach.}
  \label{fig:sup_shwang}
\end{figure*}

\begin{figure*}
  \centering
  \includegraphics[width=\linewidth]{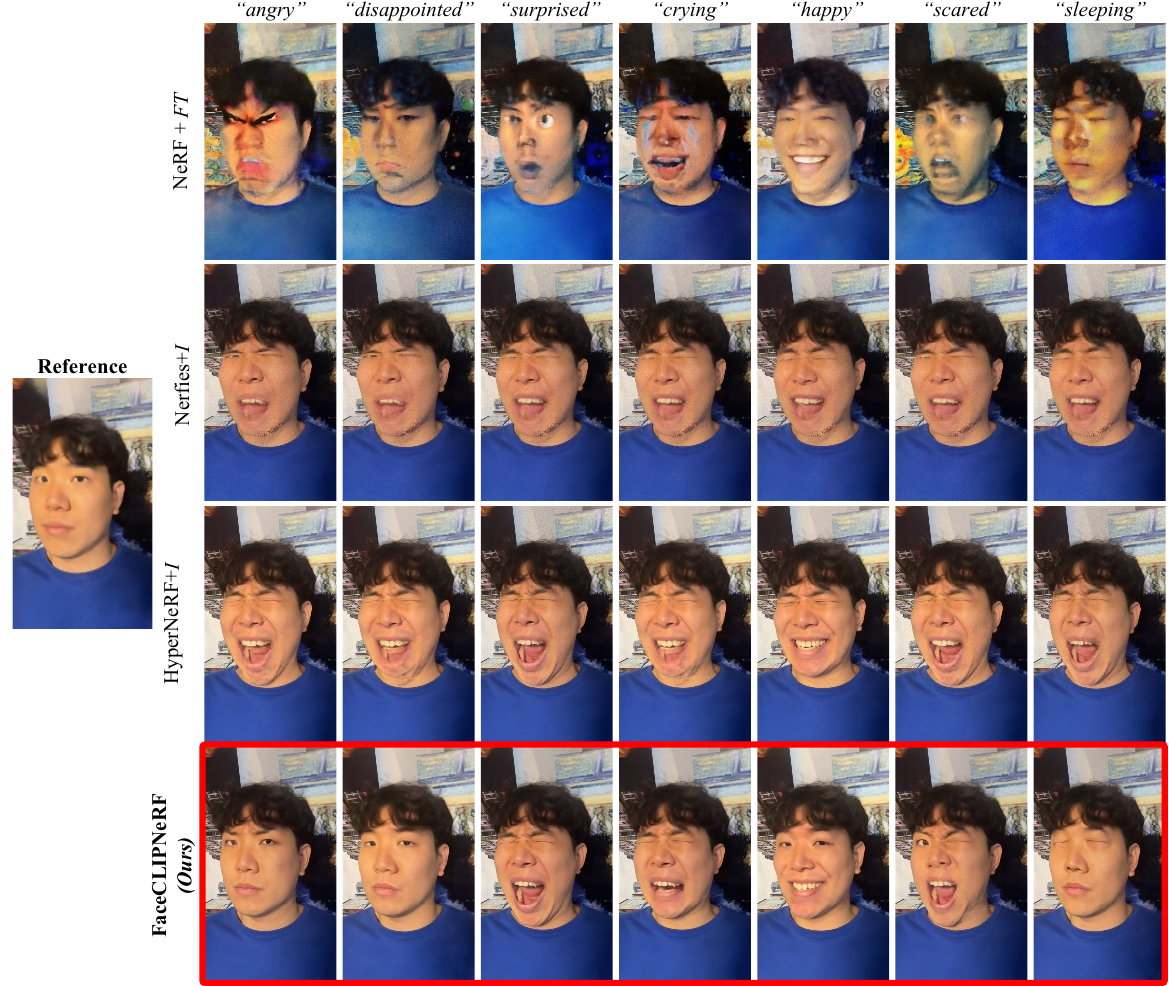}
  \caption{Text-driven manipulation results on volunteer \#6 using baseline methods and our approach.}
  \label{fig:sup_dongkim}
\end{figure*}

We report all text-driven manipulation results over all scenes captured from six volunteers in Fig \ref{fig:sup_shwang} - Fig \ref{fig:sup_dongkim}. We may also conclude from the extensive qualitative comparisons that our manipulation approach reflects target attributes most faithfully while preserving the visual quality and face identity on most results. 

\begin{figure*}
  \centering
  \includegraphics[width=\linewidth]{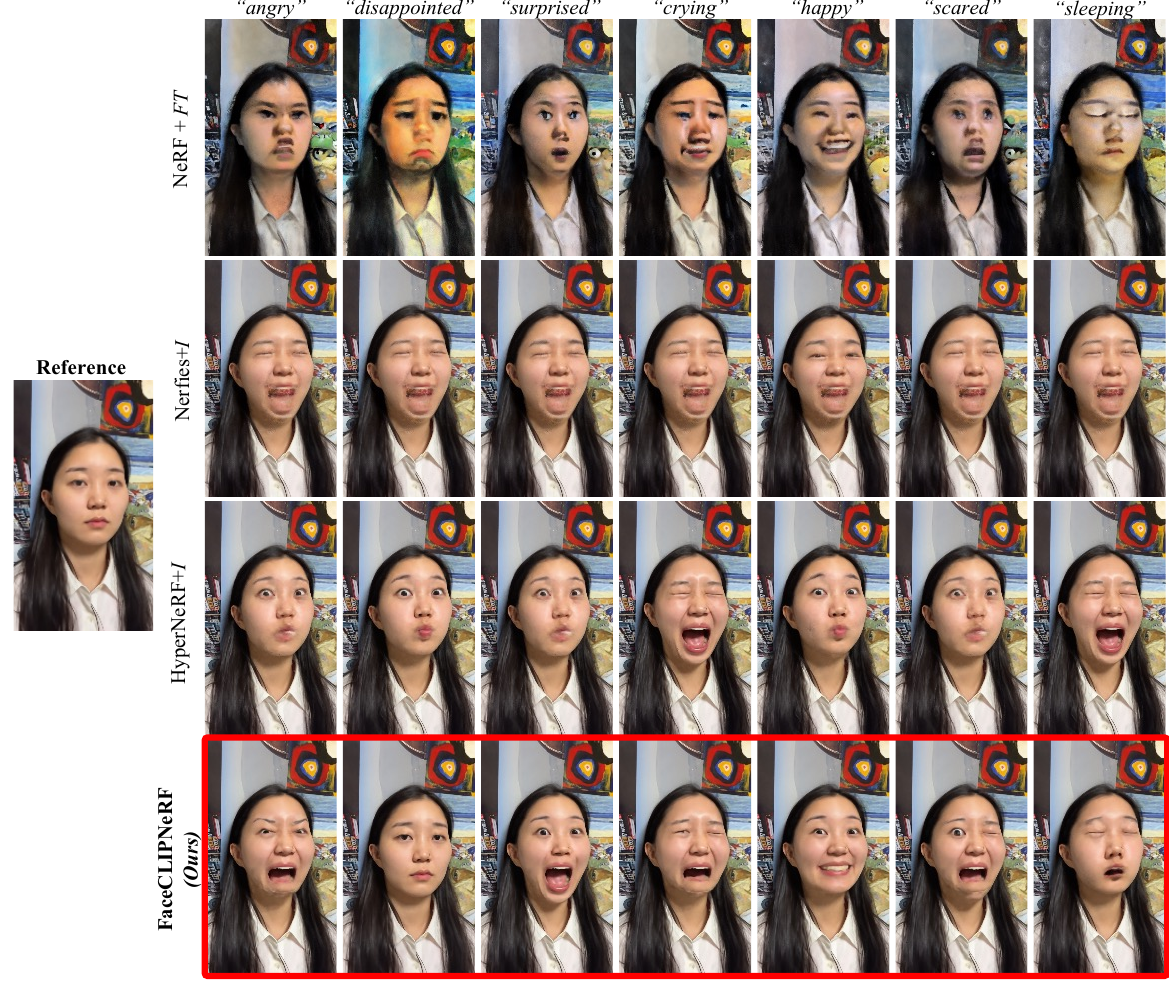}
  \caption{Text-driven manipulation results on volunteer \#5 using baseline methods and our approach.}
  \label{fig:sup_eheo}
\end{figure*}

However, there are a few failure cases to reflect on. The face manipulated with \textit{"angry"} in Fig. \ref{fig:sup_eheo} shows a face identity that is slightly different from the original, whereas the face manipulated with \textit{"crying"} in Fig. \ref{fig:sup_eheo} exhibits a noticeable degradation in visual quality. We may conclude from the observation that such a few failure cases are drawbacks of (i) Anchor Composition Network (ACN) that regularizes the natural renderings of face parts by assimilating adjacent latent codes, and of (ii) Lipschitz regularization to increase visual quality of interpolated latent codes. When ACN smoothens the adjacet latent codes, the network is not provided with prior information on facial parts, meaning that even the smoothing can be applied to unwanted locations such as boundaries of two different facial parts. Also, Lipshitz regularization is an implicit regularization method, meaning that the network is not provided with ground truth renderings of interpolated codes to be supervised. As so, further research may focus on compositing deformation given information over face priors on in order to prevent such failure cases.

\begin{figure*}
  \centering
  \includegraphics[width=\linewidth]{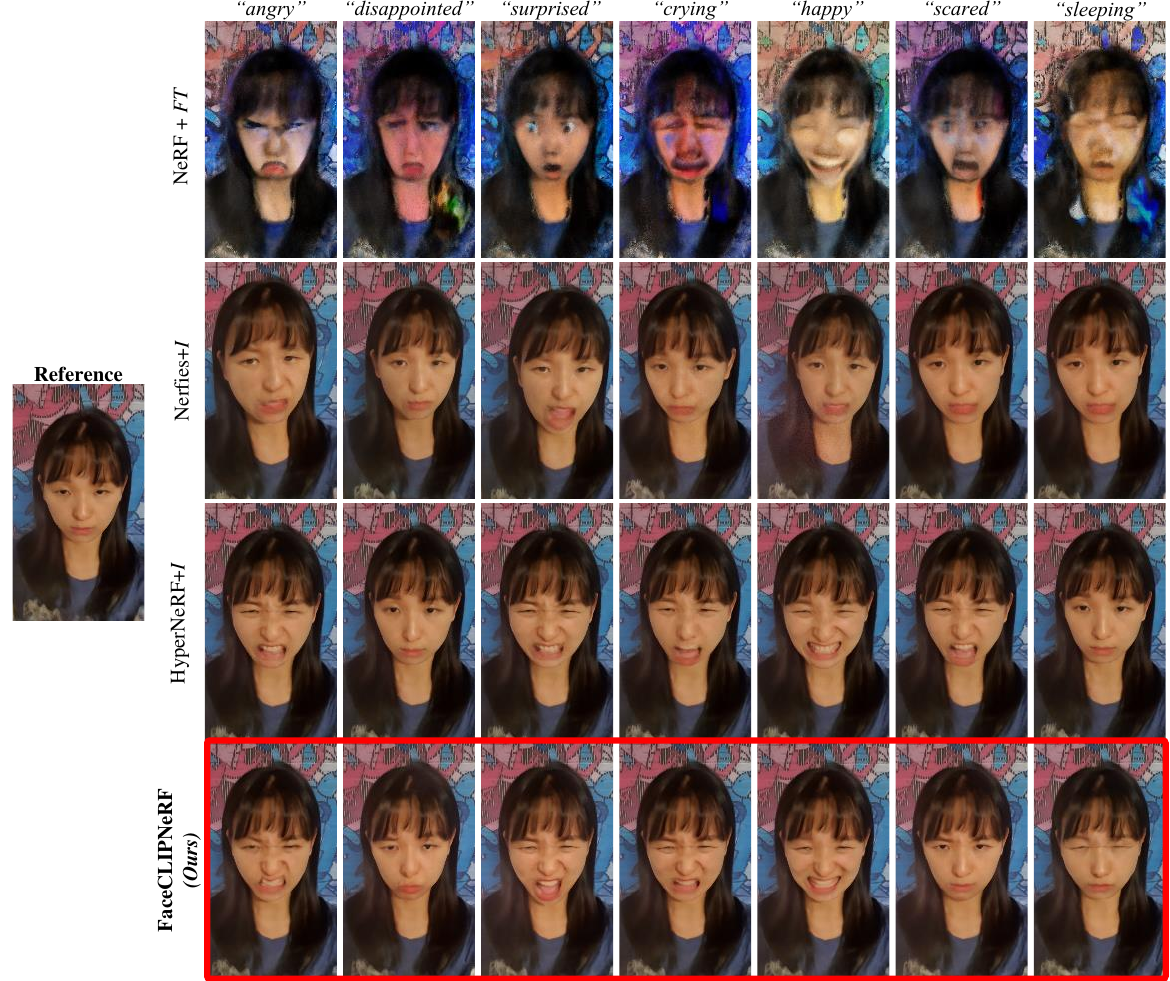}
  \caption{Text-driven manipulation results on volunteer \#2 using baseline methods and our approach.}
  \label{fig:sup_mkim}
\end{figure*}

\begin{figure*}
  \centering
  \includegraphics[width=\linewidth]{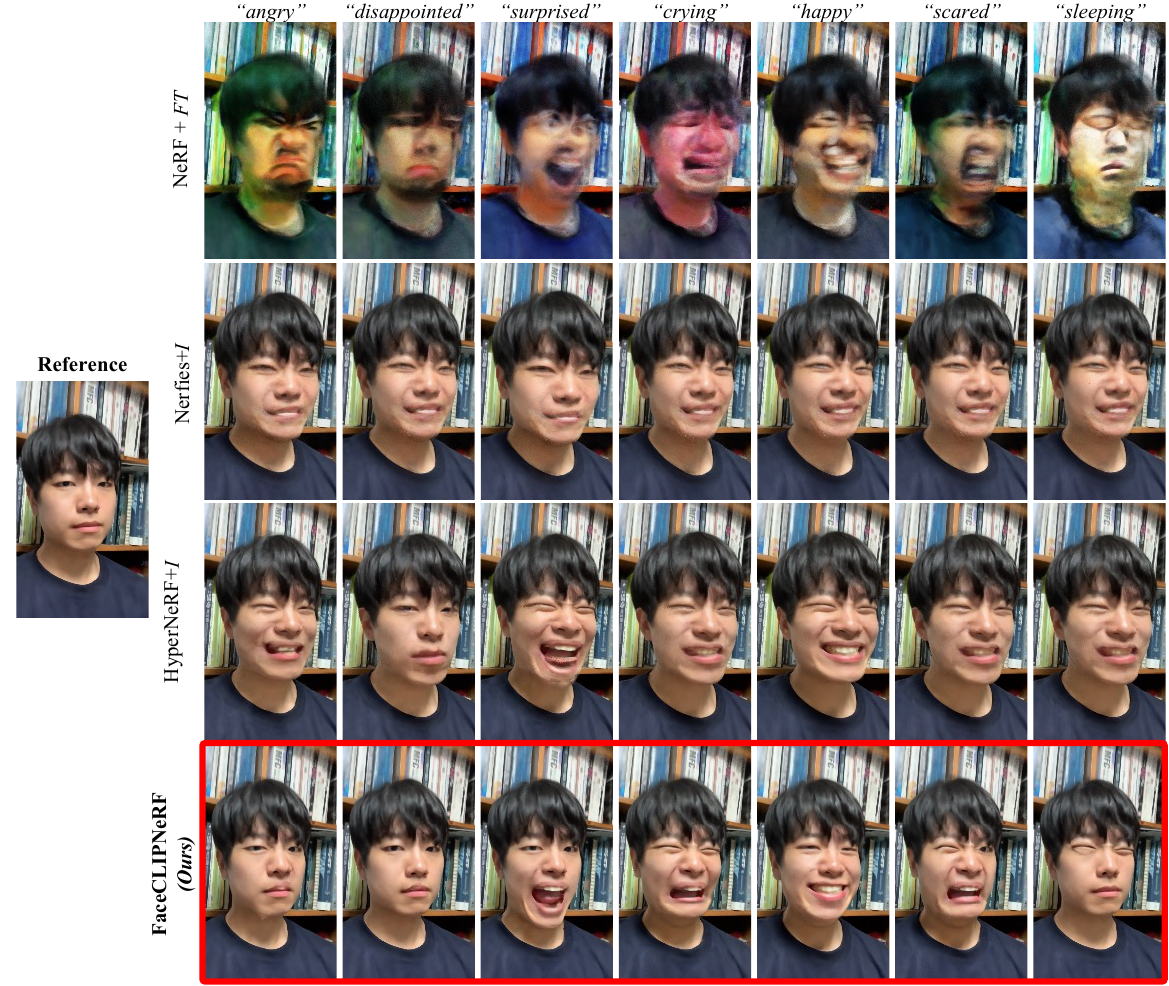}
  \caption{Text-driven manipulation results on volunteer \#3 using baseline methods and our approach.}
  \label{fig:sup_dkim}
\end{figure*}

\begin{figure*}
  \centering
  \includegraphics[width=\linewidth]{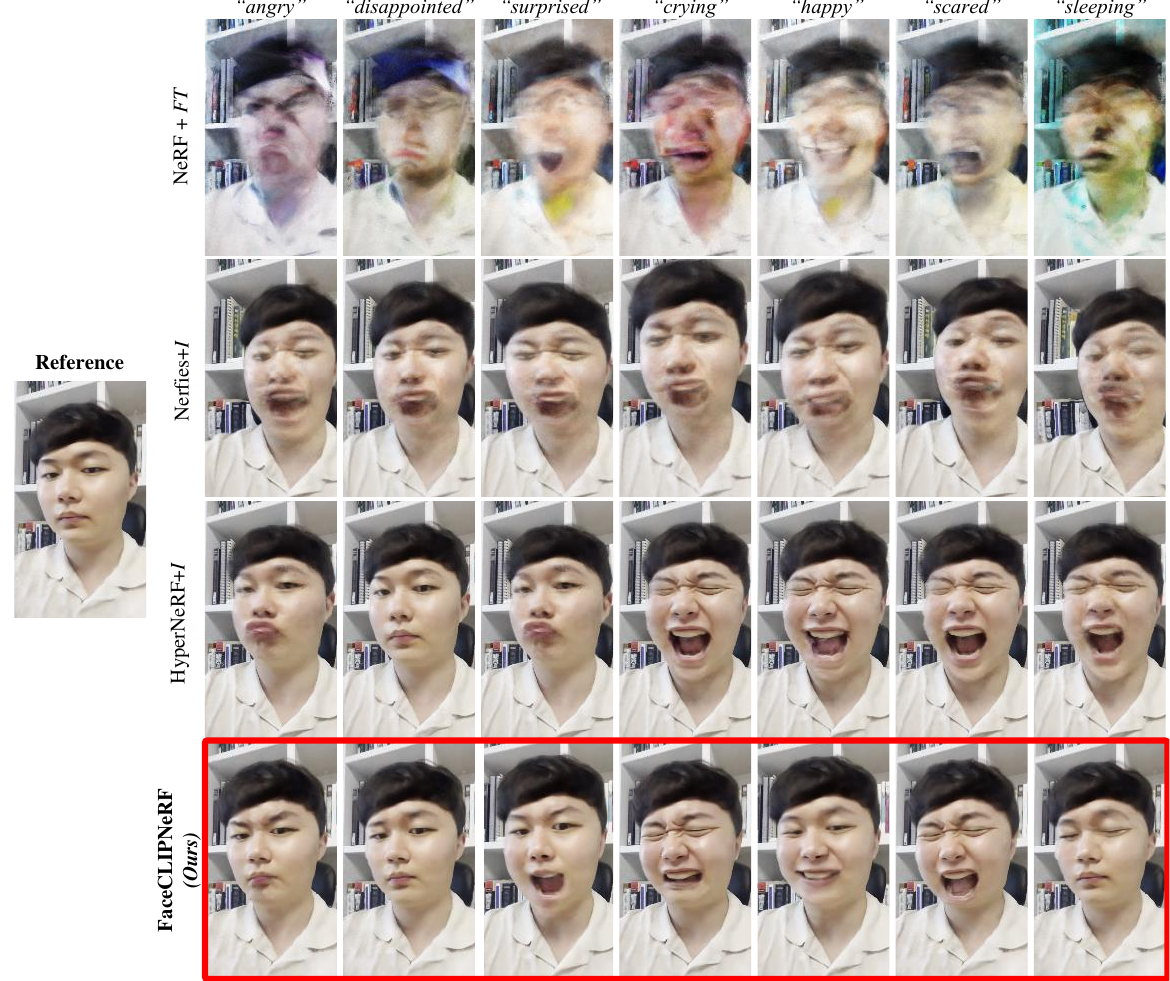}
  \caption{Text-driven manipulation results on volunteer \#4 using baseline methods and our approach.}
  \label{fig:sup_jhyung}
\end{figure*}

{\small
\bibliographystyle{ieee_fullname}
\bibliography{egbib}
}